\documentclass[10pt]{article} 

\usepackage[preprint]{rlj} 

\usepackage{amssymb}            
\usepackage{mathtools}          
\usepackage{mathrsfs}           
\usepackage{graphicx}           
\usepackage{subcaption}         
\usepackage[space]{grffile}     
\usepackage{url}                
\usepackage{lipsum}             

\usepackage{amssymb, amsmath, amsthm}
\usepackage{microtype}

\usepackage{dsfont}

\newcommand{\argmax}{\operatorname*{arg \ max}}
\newcommand{\argmin}{\operatorname*{arg \ min}}

\newcommand{\bo}{\boldsymbol}
\newcommand{\Prob}{\mathbb{P}}

\newcommand{\ru}{\rho_{\min}^{-2}U^2}
\newcommand{\lamin}{\lambda_{\min}(\bo{W}}
\newcommand{\abam}{a_t^{\textnormal{BAM}}}
\newcommand{\Abam}{A_t^{\textnormal{BAM}}}
\newcommand{\aebucb}{a_t^{\textnormal{EB-UCB}}}

\newcommand{\E}{\mathbb{E}}
\newcommand{\R}{\mathbb{R}}
\newcommand{\A}{\mathcal{A}}

\newcommand{\tn}{\textnormal}

\usepackage{etoolbox}
\AtBeginEnvironment{algorithm}{\setstretch{1.35}}

\usepackage{hyperref}
\usepackage{amsmath}
\usepackage[table]{xcolor}
\newcolumntype{L}{>$l<$}
\theoremstyle{plain}

\newtheorem{theorem}{Theorem}
\newtheorem{prop}{Proposition}

\newtheorem{assumption}{Assumption}
\newtheorem{lemma}{Lemma}

\usepackage{algorithm,algorithmic}
\usepackage{setspace}
\usepackage{tikz}

\title{Empirical Bound Information-Directed Sampling for Norm-Agnostic Bandits}

\setrunningtitle{Empirical Bound Information-Directed Sampling}

\author{Piotr M. Suder\textsuperscript{1}, Eric Laber\textsuperscript{1}}

\emails{piotr.suder@duke.edu, \ eric.laber@duke.edu}

\affiliations{
$^{1}$\textbf{Department of Statistical Science, Duke University}
}

\contribution{
    This paper introduces a novel frequentist information-directed sampling (IDS) algorithm that does not require prior knowledge of a tight upper bound of the true parameter norm to achieve good performance. Our method uses accumulating data to generate a sequence of high-probability upper bounds on the parameter norm and accounts for potential heteroskedasticity of the rewards.
    }
    {
    The performance of many frequentist bandit algorithms, including various IDS \citep[][] {kirschner2018, asymptotically_opt_ids} and UCB methods \citep[][] {auer2002_ucb, yadkori}, relies heavily on a (at least relatively) tight upper bound on the true parameter norm being available to the algorithm. This is almost never the case in practice which can lead to significant regret accumulation. Recently, some norm-agnostic bandit algorithms have been proposed to address this issue \citep{norm_agnostic}, however, they do not account for potential heteroskedasticity of the rewards.
    }
\contribution{
We introduce a new composite information criterion that balances improving the requisite upper bound on the parameter norm and direct search for the optimal action.
    }
    {
    To the best of our knowledge, no other IDS algorithm uses a mixture of information gain criteria to balance acquiring information about different aspects of the environment's dynamics. We are also not aware of any existing method that uses an information gain criterion aimed at improving the upper bound on the parameter norm.
    }

\contribution{
    We establish anytime sublinear regret bounds for our algorithm which eventually do not depend on the initially assumed parameter norm bound.
    }
    {
    Previously proposed norm-agnostic bandits \citep{norm_agnostic} rely on an initial burn-in 
during which regret accumulation is not controlled, e.g., it need not be 
sublinear.
    }

\keywords{bandit algorithms, information-directed sampling, parameter bounds, heteroskedastic noise} 

\summary{Information-directed sampling (IDS) is a powerful framework for solving bandit problems which has shown strong results in both Bayesian and frequentist settings. However, frequentist IDS, like many other bandit algorithms, requires that one have prior knowledge of a (relatively)
tight upper bound on the norm of 
the true parameter vector governing the reward model 
in order to achieve good performance. Unfortunately, this 
requirement is rarely satisfied 
in practice.  As we demonstrate, using a poorly calibrated bound can lead to significant regret accumulation. 
To address this issue, we introduce a novel frequentist IDS algorithm that iteratively refines a high-probability upper bound on the true parameter norm using 
accumulating data. We focus on the linear bandit setting with heteroskedastic subgaussian noise. Our method leverages a mixture 
of relevant information gain criteria to balance exploration aimed
at tightening the parameter norm bound and directly 
searching for the optimal action. We 
establish regret bounds for our algorithm that do not depend on an initially assumed 
parameter norm bound and demonstrate that our method outperforms 
state-of-the-art IDS and UCB algorithms.
}

\begin{document}

\maketitle  

\begin{abstract}
Information-directed sampling (IDS) is a powerful framework for solving bandit problems which has shown strong results in both Bayesian and frequentist settings. However, frequentist IDS, like many other bandit algorithms, requires that one have prior knowledge of a (relatively)
tight upper bound on the norm of 
the true parameter vector governing the reward model 
in order to achieve good performance. Unfortunately, this 
requirement is rarely satisfied 
in practice.  As we demonstrate, using a poorly calibrated bound can lead to significant regret accumulation. 
To address this issue, we introduce a novel frequentist IDS algorithm that iteratively refines a high-probability upper bound on the true parameter norm using 
accumulating data. We focus on the linear bandit setting with heteroskedastic subgaussian noise. Our method leverages a mixture 
of relevant information gain criteria to balance exploration aimed
at tightening the 
estimated parameter norm bound and directly 
searching for the optimal action. We 
establish regret bounds for our algorithm that do not depend on an initially assumed 
parameter norm bound and demonstrate that our method outperforms 
state-of-the-art IDS and UCB algorithms.

\vspace{1em}
\noindent \textbf{Keywords and phrases:} Bandit algorithms, information-directed sampling, parameter bounds, heteroskedastic noise. 

\end{abstract}

\section{Introduction}
We consider linear stochastic bandits \citep{bandit_book} with heteroskedastic noise \citep[see][for applications of such models in marketing and other areas]{weltz2023}.  In this setting, information directed sampling (IDS) and upper confidence bound (UCB) algorithms have been shown to be extremely effective \citep[][]{auer2002_ucb, yadkori, kirschner2018, asymptotically_opt_ids}.  However, many of these methods require strong prior information that can be used to inform a high-quality upper bound on the Euclidean norm of the parameter vector indexing the reward model.  The choice of this bound is critical to algorithm performance.  If the bound is too large, the algorithm risks incurring excess risk due to needless exploration, and if the bound is too small, the algorithm may fail to identify the optimal arm and thus suffer linear regret.

To reduce sensitivity on a user-specified bound, we propose a novel version of frequentist IDS that uses accumulating data to generate a sequence of high-probability upper bounds on the norm of the reward model parameters.   A key component of our method is a new information gain criterion that balances improving the requisite upper bound and regret minimization.  Because improving the bound is critical to avoid over-exploration in early rounds of the bandit process, we develop a two-phase procedure that uses our new information criterion in the first phase and then defaults to a more standard IDS information criterion in the second phase.

Unlike other bandit strategies, such as UCB \citep[][]{auer2002_ucb, kl_ucb, ucb_annals, neural_ucb} or Thompson sampling (TS) \citep[][]{thompson1933, agrawal2013thompson, approx_ts}, which encourage exploration indirectly by leveraging uncertainty about the optimal arm, IDS explicitly balances exploration and exploitation. It selects actions that minimize estimated instantaneous regret while maximizing expected information gain about model parameters. As shown by \citet{russo_ids} and \citet{kirschner2018}, this approach allows IDS to avoid pitfalls inherent in UCB and TS-based algorithms, particularly in scenarios where certain suboptimal actions provide valuable information about the environment’s dynamics. In such cases, UCB and TS tend to overlook these actions, whereas IDS plays them early on, enabling faster learning of the optimal policy and ultimately achieving superior long-term performance. IDS was first introduced for Bayesian bandits by \citet{russo_ids} and later adapted to the frequentist setting by \citet{kirschner2018}.
Beyond the standard bandit setting, IDS has been applied to problems such as linear partial monitoring \citep{partial_linear_monitoring} --- a generalization of bandits where the observed signal on the environment model parameters is not necessarily the same as the reward to be optimized --- as well as reinforcement learning \citep[][]{ids_deep_rl, ids_rl_1, ids_rl_2}, where the actions taken by the agent influence the state of the environment and the reward dynamics.

To the best of our knowledge, no previous work has considered either the strategy of iteratively refining and utilizing a high-probability upper bound on the parameter norm in the heteroskedastic subgaussian linear bandit setting we work with here, or the use of the information gain criterion for tightening the bound on the parameter norm we introduce. We are also not aware of any work utilizing a mixture of information gain criteria to encourage simultaneously obtaining different types of information about the dynamics of the environment. We note that while we introduce this idea in the form of an IDS algorithm, the approach of iteratively refining and utilizing a high-probability upper bound of the true parameter norm can be regarded as a more general design principle beyond its IDS implementation in this setting.

The remainder of this manuscript is structured as follows. The next section provides a brief review of related work. Section \ref{setup_and_notation} introduces the problem setup and notation used throughout the paper. In Section \ref{information_directed_sampling}, we present the necessary background on IDS. Section \ref{empirical_bound_ids} introduces the novel empirical bound information-directed sampling (EBIDS) algorithm, which removes the need for a tight parameter norm bound to be known {\em a priori}. Section \ref{regret_bound} establishes regret bound guarantees for EBIDS, and finally, Section \ref{simulation_study} evaluates its empirical performance against competitor algorithms in a simulation study.

\section{Related works}\label{related_works}
The assumption that the norm of the parameter indexing the reward model is known or 
that one has a (relatively) tight upper bound on this quantity is 
abundant in the IDS and UCB literature \citep[][]{auer2002_ucb, yadkori, kirschner2018, Hung_2021};
it has also been used in Thompson sampling \citep[][]{noise_adapt_TS}. 
This assumption commonly arises through the use of self-normalized 
martingale bounds and related concentration results \citep{yadkori}. 
Consequently, algorithms constructed through
these concentration results require a user-specified upper bound on the norm
or the true parameter vector.  Critically, as noted previously, the performance of these algorithms
can be highly sensitive to the choice of these bounds.  Despite  this, only a handful 
of papers have attempted to alleviate this sensitivity.  

\citet{norm_agnostic} propose norm-agnostic linear bandits which construct
a series of confidence ellipsoids for the true parameter vector along with 
a projection interval to construct a UCB-type algorithm.  However, their algorithms
rely on an initial burn-in 
during which regret accumulation is not controlled, e.g., it need not be 
sublinear.  In our simulation experiments, we find that the impact of this 
initial exploration on accumulated regret is not negligible.  Furthermore, 
as UCB algorithms, their methods do not explicitly make use of heteroskedasticity 
in the reward distributions across arms.

The algorithm proposed by \citet{ghosh21} shares some underlying ideas with our method in the sense that they use multi-phase exploration to iteratively update the bound on the unknown parameter norm. However, their algorithm is limited to the specialized setting of stochastic linear bandits introduced by \citet{stoch_bandit} with restrictive assumptions on the structure of the rewards which 
makes their methods generally not applicable to the settings we consider here.  
Similarly, \citet{dani_2008}, \citet{orabona2011}, and \citet{gentile14} do not assume 
that one has a high-quality (i.e., relatively tight) bound on the norm of 
the parameter; however, they require bounded rewards for all arms. Other attempts to alleviate the assumption of known parameter norm bound have been made in spectral bandits \citep{spectral_bandits}, and deep active learning \citep{wang_2021}. However, it is not clear how to port these methods to
the setup we consider here.  



\section{Setup and notation}\label{setup_and_notation}
We denote the inner product of two vectors of the same dimension as $\langle \cdot , \cdot \rangle$
so that the squared Euclidean norm of vector $\bo{v}$ is $||\bo{v}||_2^2 = 
{\langle \bo{v}, \bo{v}\rangle}$.  
For a symmetric positive definite or semi-definite matrix $\bo{A} \in \mathbb{R}^{d \times d}$, we denote the associated matrix norm (or semi-norm) of a vector $\bo{v} \in \mathbb{R}^d$ as $\| \bo{v}\|_{\bo{A}}^2 = \langle \bo{v}, \bo{A}\bo{v} \rangle$. 
We let $\lambda_{\max}(\bo{A})$ and $\lambda_{\min}(\bo{A})$ 
denote the largest and the smallest eigenvalues of $\bo{A}$. Throughout, $\log(x)$ denotes
the natural logarithm of $x\in\mathbb{R}_+$. 

At each time step $t \in \{1, \ldots , T\}$ 
the agent selects an action $A_t\in\mathcal{A}$ and observes the outcome 
$Y_t \in \R$ which is generated from the linear model 
\begin{equation}\label{eq:reward_model}
Y_t(A_t) = \langle \bo{\phi}(A_t), \bo{\theta}^* \rangle + \eta_t,
\end{equation}
where $\bo{\theta}^* \in \R^d$ is a vector of unknown parameters and $\bo{\phi}: \mathcal{A} \rightarrow \mathbb{R}^d$ is a feature mapping, such that for any $a \in \A$ 
we have $\| \bo{\phi}(a) \|_2 \in [L, U]$ for some positive constants $L \leq U$. 
The noise term $\eta_t$ is assumed to be subgaussian and conditionally mean zero, i.e., for
every $c \in \R$  we assume that  
\begin{equation}\label{eqn:subgaussian}
\mathbb{E}
\left\lbrace 
    \exp \left(c \eta_t\right) \mid A_t = a
\right\rbrace  
\leq 
\exp \left\lbrace 
c^2 \rho\left(a\right)^2 / 2
\right\rbrace, 
\end{equation}
where $0 < \rho_{\min} \le \rho(a) \le \rho_{\max} < \infty$ for all $a\in\A$ 
and 
$\E\left( 
\eta_t \mid A_1, \ldots , A_t, \eta_1, \ldots , \eta_{t-1}
\right) = 0.$
Define $B^* := \| \bo{\theta}^* \|_2$, 
in some of our theoretical results we assume that one has a conservative upper bound  
$B$ such that $B^* \le B$ but that this bound may be quite conservative, 
i.e., it may be that  $B^* \ll B$.

The available history to inform action selection at time $t$ is 
$\bo{H}_t = \{(A_1, Y_1), \ldots , (A_{t-1}, Y_{t-1})\}$ of past actions and rewards. 
A bandit algorithm is thus formalized as a map from histories to distributions over
actions 
$\pi_t(a| \bo{h}_t) = \mathbb{P}(A_t =a|\bo{H}_t = \bo{h}_t)$.  
Let
$$\Delta(A_t) = \langle \bo{\phi}(a^*), \bo{\theta}^* \rangle - \langle \bo{\phi}(A_t), \bo{\theta}^* \rangle $$
be the gap between the action $A_t$ and the optimal action $a^* = \argmax_{a \in \A} \langle \bo{\phi}(a), \bo{\theta}^* \rangle$. 
Our goal is to design an algorithm $\pi_t(\cdot \mid \bo{h}_t)$ which maximizes the cumulative expected reward $\E \left\lbrace 
\sum_{t=1}^{T} Y_t \right\rbrace$, 
or equivalently, minimizes the {regret}, defined as 
$
\mathcal{R}_T = \E \left\lbrace \sum_{t=1}^{T} \Delta(A_t) \right\rbrace.
$
While regret is a standard performance metric for bandit algorithms, 
it involves taking expectation over both the randomness in the policy and the noise in the rewards so it can 
be a poor indicator of the risk associated with the policy \citep{bandit_book}. For this reason in this paper 
we also study the probabilistic bounds on the {pseudo-regret} defined as
$
\mathcal{PR}_T = \sum_{t=1}^{T} \Delta(A_t).
$

\section{Review of information-directed sampling}\label{information_directed_sampling}
Information-directed sampling \citep[IDS][]{russo_ids} is an
algorithm design principle that balances minimizing the
gap of an action with its potential for information gain.  
Let $\mathcal{P}(\A)$ denote the space of distributions 
of $\A$.  For any $\mu \in \mathcal{\A}$ let 
$\widehat{\Delta}_t(\mu)$ be an estimator of the expected
gap $\mathbb{E}_{\mu}\Delta := 
\mathbb{E}_{A\sim \mu}\Delta(A)$ constructed from the history $\bo{H}_t$, 
and, similarly, let 
$I_t(\mu)$ be a measure of information again, e.g., 
the reduction of entropy in the posterior or sampling
distribution the parameter indexing the mean 
reward model (see below for additional details).   
For any function $f:\mathcal{P}(\mathcal{A})\rightarrow \mathbb{R}$, 
if the argument is a point mass 
at a single action, e.g., where $\mu$ is the Dirac delta 
$\delta_a$, we write $f(a)$ rather than
$f(\delta_a)$.  
The IDS distribution is defined as 
\begin{equation}\label{eq:ids}
\mu_t^{\text{IDS}} = \argmin_{\mu \in \mathcal{P}(\A)}
\frac{\left\lbrace \widehat{\Delta}_{t}(\mu)\right\rbrace^2}{I_t(\mu)}. 
\end{equation}
 The quantity $\Psi_t(\mu) := 
 {\left\lbrace \widehat{\Delta}_{t}(\mu)\right\rbrace^2}/{I_t(\mu)}$ 
 being minimized is known as the \textit{information ratio}. An IDS algorithm samples the action $A_t \sim \mu_t^{\text{IDS}}$ at each time step $t$.
Note that this results in a randomized algorithm, which, as shown by \citet{russo_ids} and \citet{kirschner2018}, always has at most two actions in its support. However, it is also possible to restrict the optimization in \eqref{eq:ids} to Dirac delta functions on the individual actions, thus obtaining what is often referred to as \textit{deterministic IDS} \citep{kirschner2018} 
\begin{equation}\label{eq:dids}
\widehat{A}_t^{\text{DIDS}} = \argmin_{a \in \A}
\frac{\left\lbrace \widehat{\Delta}_{t}(a)\right\rbrace^2}{I_t(a)}.
\end{equation}
Deterministic IDS is typically computationally cheaper, 
retains the same theoretical regret bounds as its randomized counterpart, and in simulation
experiments 
was shown to be competitive with
or superior to randomized IDS 
\citep{kirschner2018, kirschner_thesis}. Furthermore, deterministic IDS may be appealing 
in settings where randomized policies are unpalatable such as public health
\citep[][]{weltz2022}
and site selection \citep[][]{ahmadi2017survey}.

The information ratio provides a natural way of bounding regret within a Bayesian setting
\citep[][]{russo_ids}. 
Notably, the information ratio can also be used to bound the regret under
a frequentist paradigm \citep[][]{kirschner2018} as illustrated by the
following result based on the work of \citet[][]{kirschner_thesis} which we prove in  Section \ref{proof_thm_1} of  
the Supplementary Materials.  
\begin{theorem}[Kirschner]\label{thm:univ_IDS_bound}
For any $T$ let $G$ be a fixed subset of $\{1, \ldots , T\}$ and 
let $\{A_t\}_{t = 1}^T$ be an $\bo{H}_t$-adapted sequence in $\A$. Then
$$
\mathbb{E}
\left\lbrace 
    \sum_{t \in G} \widehat{\Delta}_t \left(A_t\right)
\right\rbrace \leq 
\sqrt{ \E\left\lbrace 
\sum_{t \in G} \Psi_t\left(A_t \right)
\right\rbrace 
\E\left\lbrace 
\sum_{t \in G} I_t\left(A_t\right)
\right\rbrace},
$$
and if $\widehat{\Delta}_t(A_t) \geq \Delta(A_t)$ for all $t \in G $
then with probability $1$ we have
$$ \sum_{t \in G} {\Delta}\left(A_t\right) \leq 
\sqrt{
\left\lbrace 
    \sum_{t \in G}\Psi_t(A_t)
\right\rbrace  
\left\lbrace 
    \sum_{t \in G} I_t(A_t)
\right\rbrace 
}.$$
\end{theorem}

\citet{kirschner2018} used weighted ridge regression to estimate $\bo{\theta}^*$ 
at each time step $t$ so that 
\begin{equation}\label{eq:wls_def}
    {\widehat{\bo{\theta}}}^{\mathrm{wls}}_t = 
    \bo{W}_t^{-1} \sum_{s=1}^{t-1} \frac{1}{\rho(A_s)^2}\bo{\phi}(A_s) Y_s, \quad \text{ where } \quad \bo{W}_t = \sum_{s=1}^{t-1}  \frac{1}{\rho(A_s)^2}\bo{\phi}(A_s)\bo{\phi}(A_s)^{\top} + \gamma \bo{I}_d,
\end{equation}
and $\gamma \ge 0$ is a constant chosen by the user. 
The following result, proposed by \citet{yadkori} and extended by \citet{kirschner2018}, 
provides a means to perform inference using this estimator.

\begin{theorem}\label{thm:snmb}
Suppose that the generative model follows the linear bandit model
$Y_t=\left\langle \bo{\phi}(A_t), \bo{\theta}^*\right\rangle+\eta_t$
given in (\ref{eq:reward_model}), where the actions 
$A_t$ are $\bo{H}_t$-adapted and  
the errors $\eta_t$ have 
conditional mean of zero and satisfy 
the subgaussian condition in (\ref{eqn:subgaussian}).
Let $B \ge ||\bo{\theta}^*||_2$ be a (potentially conservative) bound
on the norm of the parameters indexing the reward model and define 
\begin{equation*}
\mathcal{E}_{t, \delta}^{\mathrm{wls}} := \left\{\bo{\theta} \in \mathbb{R}^d:\left\|\bo{\theta}-{\widehat{\bo{\theta}}}_t^{\mathrm{wls}}\right\|_{\bo{W}_t}^2 \leq \beta_{t, \delta}(B) \right\},
\end{equation*}
where 
\begin{equation}\label{eq:beta_def}
\beta_{t, \delta}(B)=\left[
\sqrt{2 \log \frac{1}{\delta}+\log 
\left\lbrace 
\frac{\operatorname{det}\left(\bo{W}_t\right)}{\operatorname{det}\left(\bo{W}_1\right)}
\right\rbrace}+\sqrt{\gamma} B\right]^2.
\end{equation}

Then
$$
\mathbb{P}\left(\bigcap_{t=1}^{\infty}\left\{\bo{\theta}^* \in \mathcal{E}_{t, \delta}^{\mathrm{wls}} \right\}\right) \geq 1-\delta,
$$
i.e., $\mathcal{E}_{t, \delta}^{\mathrm{wls}}$ is a $(1-\delta)\times 100\%$ 
confidence ellipsoid for $\bo{\theta^*}$.
\end{theorem}

\citet{kirschner2018} use Theorem \ref{thm:snmb} to formulate a weighted UCB algorithm which at each time step $t$ takes the action 
\begin{equation}\label{eq:UCB_action}
A_t^{\tn{UCB}(\delta_t)} = \argmax_{a \in \A} \left\langle \bo{\phi}(a), \widehat{\bo{\theta}}_t^{\mathrm{wls}}\right\rangle + 
\beta^{1/2}_{t, \delta_t}(B)\| \bo{\phi}(a)\|_{\bo{W}_t^{-1}},
\end{equation}
maximizing the $(1-\delta_t)\times 100\%$ upper confidence bound 
on the expected reward based on the 
$\mathcal{E}_{t, \delta_t}^{\mathrm{wls}}$ confidence set. Then they use
\begin{equation*}
    \check{\Delta}_{t, \delta_t}(a) = \left\langle \bo{\phi}\left(A_t^{\tn{UCB}(\delta_t)}\right) - \phi(a), \widehat{\bo{\theta}}_t^{\mathrm{wls}}\right\rangle+{\beta}_{t, \delta_t}^{1 / 2}(B) \left(\left\|\bo{\phi}\left(A_t^{\tn{UCB}(\delta_t)}\right)\right\|_{\bo{W}_t^{-1}} + \|\bo{\phi}(a)\|_{\bo{W}_t^{-1}} \right).
\end{equation*}
as the gap estimate.
This ensures that $\Delta(a) \leq \check{\Delta}_{t, \delta_t}(a)$ for all $a \in \A$ whenever $\bo{\theta}^* \in \mathcal{E}_{t, \delta_t}^{\mathrm{wls}}$ holds.

 The choice of the information gain criterion is crucial when designing an IDS algorithm. \citet{kirschner2018} introduce the following criterion
\begin{align*}
 I_{t}^{\text{UCB}(\delta_t)}(a) =& \frac{1}{2} \log \left( \frac{{\left\| \bo{\phi} \left(a_t{^{\text{UCB}(\delta_t)}}\right) \right\|_{\bo{W}_t^{-1}}^2}}{{ \left\| \bo{\phi} \left(a_{t}^{\text{UCB}(\delta_t)}\right) \right \|_{(\bo{W}_t + \rho(a)^{-2} \bo{\phi}(a) \bo{\phi}(a)^{\top})^{-1}}^2}}\right),
\end{align*}
for any $a \in \A$.
We present the resulting procedure in Algorithm \ref{algo:ids_ucb}, which we 
hereafter refer to as IDS-UCB. 
\begin{algorithm}[H]\caption{IDS-UCB} \label{algo:ids_ucb}
   \setstretch{1.0}
   Input: Action set $\mathcal{A}$, penalty parameter $\gamma>0$, noise function $\rho: \mathcal{A} \rightarrow \mathbb{R}_{+}$, feature function $\bo{\phi}: \A \rightarrow \mathbb{R^d}$, sequence of confidence levels $\{\delta_t\}_{t \geq 1} \subset (0,1)$, assumed true parameter norm bound $B$.
~~~~
\vspace{2pt}
\begin{itemize}
    \setlength{\itemindent}{-10pt}
    \item[] For $t = 1,2, \ldots, T:$
    \begin{itemize}
        \setlength{\itemindent}{-10pt}
        \item[] Compute $\bo{W}_t$ and $\widehat{\bo{\theta}}^{\mathrm{wls}}_t$ using \eqref{eq:wls_def}
        \item[] $A_t^{\tn{UCB}(\delta_t)} \leftarrow \argmax_{a \in A} \left\{ \left\langle \bo{\phi}(a), \widehat{\bo{\theta}}_t^{\mathrm{wls}}
        \right\rangle + {\beta}_{t, \delta_t}^{1 / 2}(B)\|\bo{\phi}(a)\|_{\bo{W}_t^{-1}} \right\} $
         \item[]  $I_t^{\text{UCB}(\delta_t)}(a) \leftarrow \frac{1}{2} \log \left( {{\left\| \bo{\phi} \left(A_t^{{\text{UCB}(\delta_t)}}\right) \right\|_{\bo{W}_t^{-1}}^2}}\right) - \frac{1}{2} \log \left({{ \left\| \bo{\phi} \left(A_{t}^{\text{UCB}(\delta_t)}\right) \right \|_{(\bo{W}_t + \rho(a)^{-2} \bo{\phi}(a) \bo{\phi}(a)^{\top})^{-1}}^2}}\right)$
          \item[] ${\check{\Delta}}_{t, \delta_t}(a) \leftarrow \left\langle \bo{\phi}\left(A_t^{\tn{UCB}(\delta_t)}\right) - \phi(a), \widehat{\bo{\theta}}_t^{\mathrm{wls}}\right\rangle+{\beta}_{t, \delta_t}^{1 / 2}(B) \left(\left\|\bo{\phi}\left(A_t^{\tn{UCB}(\delta_t)}\right)\right\|_{\bo{W}_t^{-1}} + \|\bo{\phi}(a)\|_{\bo{W}_t^{-1}} \right)$
        \item[] $\mu_t \leftarrow \argmin_{\mu \in \mathcal{P}(\mathcal{A})} {\check{\Delta}^2_{t, \delta_t}\left(\mu_t\right)}/{I_t^{\text{UCB}(\delta_t)}\left(\mu_t\right)}$
        \item[] Sample $A_t \sim \mu_t$
        \item[] Play $A_t$, observe $Y_t  = \left\langle \bo{\phi}(A_t), \bo{\theta}^*\right\rangle+\eta_t$
    \end{itemize}

\end{itemize}

\end{algorithm}
It can be shown that if one chooses 
$\delta_t = 1/t^2$, 
the regret of IDS-UCB satisfies 
\begin{equation*}
    \mathcal{R}_T \leq 
    O\left(\max\{U/\sqrt{\gamma}, \rho_{\max}\}\sqrt{\gamma}dB\sqrt{T} \log T\right),
\end{equation*}
while the pseudo-regret $\mathcal{PR}_T$ of IDS-UCB with fixed $\delta_t = \delta$ satisfies with probability at least $1-\delta$
$$
\mathcal{PR}_T \leq O(\max\{U/\sqrt{\gamma}, \rho_{\max}\}\sqrt{\gamma}dB\sqrt{T} \log (T/\delta));
$$
critically, 
both regret bounds scale directly with the assumed bound $B$ on the Euclidean norm of the true parameter \citep[see][for a formal statement of the preceding results and additional discussion]{kirschner_thesis}. 

We now demonstrate via a simple illustrative 
simulation experiment that the choice of $B$ can have a significant 
impact on the finite time performance of IDS-UCB.  Large values of $B$ relative to 
$B^*$ lead to excess exploration and large regret in early rounds of the algorithm, whereas
small values of $B$ can prevent the algorithm from identifying the optimal arm thus 
incurring linear regret.  
In this experiment we also include the weighted UCB policy given by 
\eqref{eq:UCB_action}. We evaluate versions of IDS-UCB and UCB that use a conservative value of 
$B > B^*$, and those which use an anti-conservative value $B < B^*$. The 
parameters indexing the generative model are $\bo{\theta}^* = 
[-5, 1, 1, 1.5, 2]^{\top}$ 
so that $B^* = \| \bo{\theta}^* \|_2 \approx 5.77$. We take $B=100$ for the 
conservative bound and $B=1$ for the anti-conservative bound. 
For reference, we also include  \textit{oracle} versions of UCB and IDS-UCB that have access to the true value of $B^*$. However, we emphasize that these procedures are not generally possible in practice. 

\begin{figure}[htbp]
    \centering
    \begin{subfigure}[b]{0.41\textwidth}
        \includegraphics[scale=0.29]{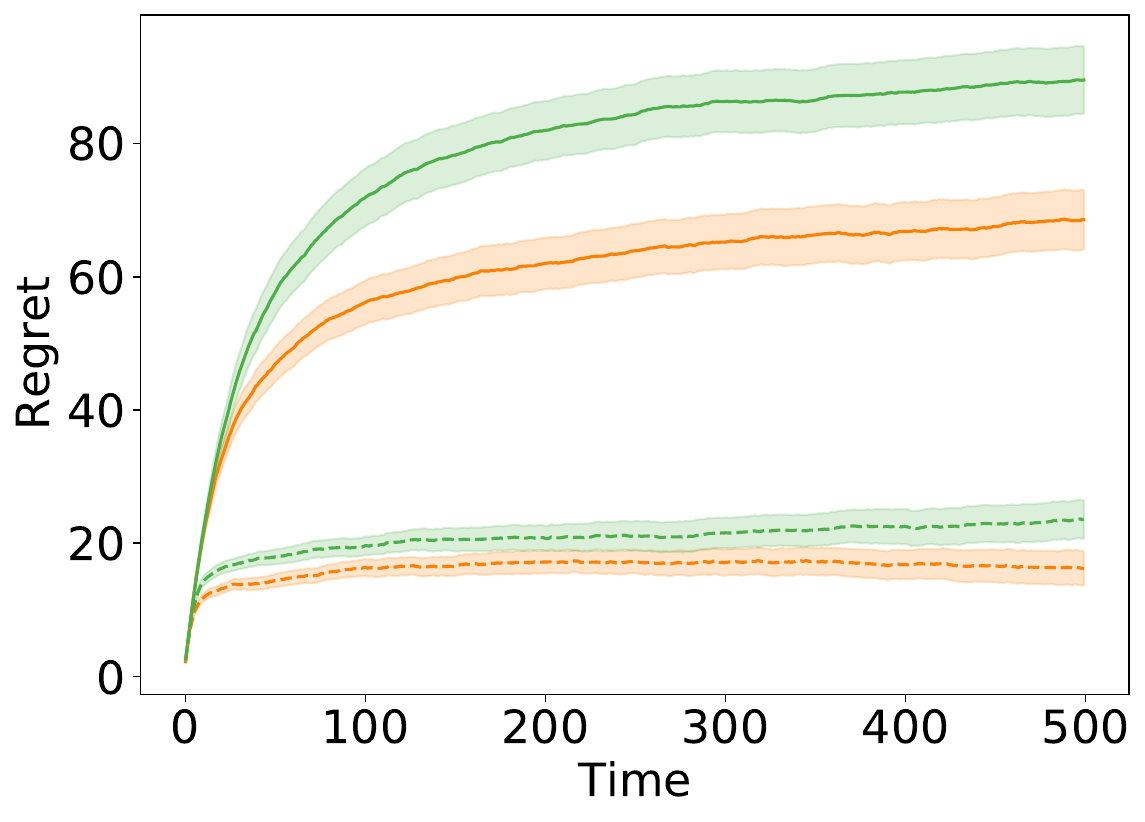}
        \caption{Conservative $B = 100$}
        \label{fig:sub1}
    \end{subfigure}
    \hfill
    \begin{subfigure}[b]{0.58\textwidth}
        \includegraphics[scale=0.29]{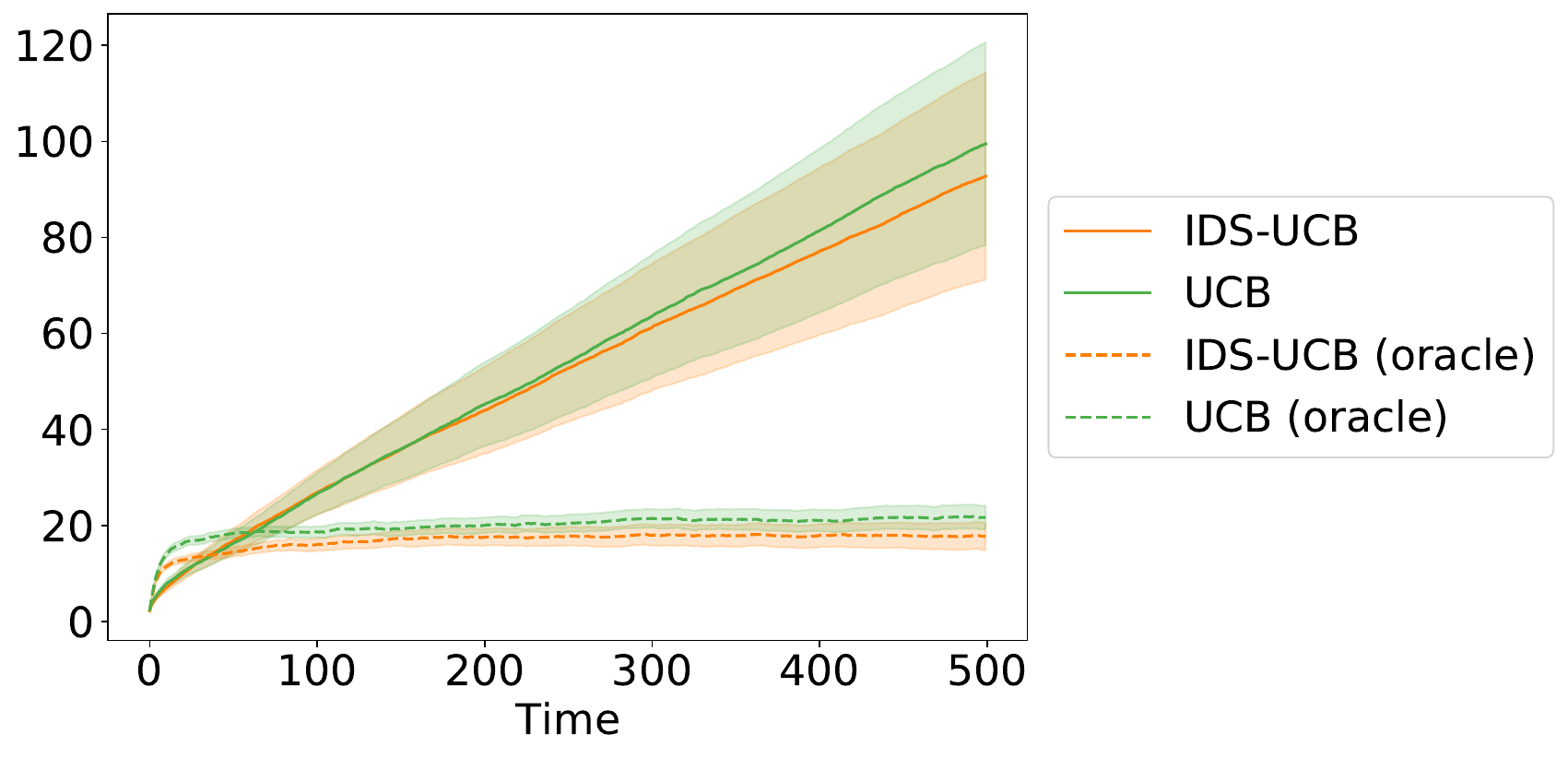}
        \caption{Anti-conservative $B = 1$ \phantom{aaaaaaaaa}}
        \label{fig:sub2}
    \end{subfigure}
    \caption{Regret incurred by IDS-UCB and UCB with: (a) conservative $B=100$; (b) anti-conservative $B=1$. In both plots we include the oracle versions of IDS-UCB, and UCB using $B=B^*$ for reference. However, note that it is not feasible to implement them in most practical settings. The solid and dashes lines represent the regret averaged over $200$ repeated experiments, while the shaded bounds are $95\%$ pointwise confidence bands. }
    \label{fig:failures}
\end{figure}

We consider a setting with ten arms. 
Features for each arm are sampled 
from  Unif$[-1/\sqrt{5}, 1/\sqrt{5}]$.  The error distribution for the first five arms
are standard normal and for the remaining five arms they are normal with mean zero and
standard deviation $0.2$.  
Figure \ref{fig:failures} shows the mean regret averaged over $200$ repeated experiments with $T=500$ steps
along with $95\%$ pointwise confidence bounds. 
As anticipated, using a conservative bound of $B=100$ achieves sublinear regret but pays 
a strong initial cost due to excess exploration.  
Algorithms that used the anti-conservative bound of $B=1$ fail to identify the optimal arm thus
sustain linear regret.

\section{Empirical bound information-directed sampling}\label{empirical_bound_ids}
We propose the empirical bound information-directed sampling (EBIDS) 
algorithm, which, like existing IDS algorithms, relies on a conservative 
upper bound $B$, but, unlike existing algorithms,
EBIDS refines this value with accruing data 
to obtain a tighter high-probability bound
on $B^*$.  
Our algorithm proceeds in two phases. 
Throughout the first $T_B$ steps, which we will refer to as the \textit{bound exploration phase}, the goal is to gather initial information on the optimal action as well as to improve the 
bound on $B^*$. At each time step $t$ in this first phase, we use
\begin{equation}\label{eq:hat_B_def}
    \widehat{B}_t = \min \left\{ B, \| \widehat{\bo{\theta}}_{t}^{\mathrm{wls}} \|_2 + \beta_{t, \zeta_t(\delta)}^{1/2}(B) \lambda_{\min}(\bo{W}_{t})^{-1/2} \right\}.
\end{equation}
as the upper bound on $B^*$.
 The term $\beta_{t,\zeta_t(\delta)}(B)$ is defined in \eqref{eq:beta_def} and $\zeta_t(\delta) = 
 \min \{\delta, 1/t^2\}$, where $\delta > 0$ is a user-specified parameter that determines the 
 confidence level for the upper bound on $B^*$. The geometric motivation for this estimator stems from the 
 fact that the confidence set ${\mathcal{E}}_{t, \zeta_t(\delta)}^{\mathrm{wls}}$ is an ellipsoid centered at 
 $\widehat{\bo{\theta}}^{\mathrm{wls}}_t$ 
 with the longest semi-axis of length $\beta_{t, \zeta_t(\delta)}^{1/2}(B) \lambda_{\min}(\bo{W}_{t})^{-1/2}$, 
 so by adding it to 
 $\| \widehat{\bo{\theta}}^{\mathrm{wls}}_t \|_2$, by the triangle inequality, we obtain a 
 conservative upper bound on the distance between the origin and the point of 
 ${\mathcal{E}}_{t, \zeta_t(\delta)}^{\mathrm{wls}}$ furthest from it. We prove in the 
 Supplementary Materials that 
$$\mathbb{P}\left(\bigcap_{t=1}^{\infty}\left\{\widehat{B}_t \geq B^* \right\}\right) \geq 1-\delta.$$

Continuing our description of the bound exploration phase, 
for any $t\leq T_B$ we use $\widehat{B}_t$ to obtain a UCB algorithm, 
which we will refer to as empirical bound UCB (EB-UCB) via
\begin{equation}\label{eq:def_EB_UCB}
A_t^{\textnormal{EB-UCB}(\zeta_t(\delta))} =  \argmax_{a \in \A} \left\langle \bo{\phi}(a), \widehat{\bo{\theta}}_t^{\mathrm{wls}}\right\rangle+\beta_{t, \zeta_t(\delta)}^{1 / 2}(\widehat{B}_t)\|\bo{\phi}(a)\|_{\bo{W}_t^{-1}}.
\end{equation}
Subsequently, we use 
\begin{align}\label{eq:def_Delta_hat}
\widehat{\Delta}_{t, \zeta_t(\delta)}(a) =& \left\langle \bo{\phi}\left(A_t^{\tn{EB-UCB}(\zeta_t(\delta))}\right) - \bo{\phi}(a), \widehat{\bo{\theta}}_t^{\mathrm{wls}}\right\rangle \nonumber \\
&+ \beta_{t, \zeta_t(\delta)}^{1 / 2}(\widehat{B}_t) \left(\left\|\bo{\phi}\left(A_t^{\tn{EB-UCB}(\zeta_t(\delta))}\right)\right\|_{\bo{W}_t^{-1}} + \|\bo{\phi}(a)\|_{\bo{W}_t^{-1}}\right)
\end{align}
as the gap estimate for any $a \in \A$. We define a new information gain
criterion that combines model improvement (classic information gain) with
bound improvement.  
The first component of our new information gain criterion is 
given by 
\begin{equation}\label{eq:def_info_EB_UCB}
    I_t^{\textnormal{EB-UCB}(\zeta_t(\delta))}(a) =  \frac{1}{2} \log \left( \frac{\left\| \bo{\phi} \left(A_t^{\text{EB-UCB}(\zeta_t(\delta))}\right) \right \|_{\bo{W}_t^{-1}}^2}{ \left\| \bo{\phi} \left(A_t^{\text{EB-UCB}(\zeta_t(\delta))}\right) \right\|_{(\bo{W}_t + \rho(a)^{-2} \bo{\phi}(a) \bo{\phi}(a)^{\top})^{-1}}^2}\right),
\end{equation}
for any $a \in \A$. It can be seen that this is analogous to the IDS-UCB information gain 
criterion considered by \citet{kirschner2018}. 
To ensure that improvement in the bound on $B^*$, we introduce the second 
component of our information gain criterion $I_t^B$ which is given by 
$$
I_t^B(a) = \frac{1}{2}\log \left( \| \bo{v}_t^{\min} \|_{(\bo{W}_t + \rho(a)^{-2} \bo{\phi}(a) \bo{\phi}(a)^{\top})}^2\right) - \frac{1}{2}\log \left\lbrace
\lambda_{\min}(\bo{W}_t) \right\rbrace,
$$
where $\bo{v}_t^{\min}$ is the unit-length eigenvector of $\bo{W}_t$ associated with the smallest eigenvalue $\lambda_{\min}(\bo{W}_t)$.  
The maximizer of $I_t^B(a)$ corresponds to the feature
vector $\phi(a)$ which generates the most (weighted) information
in the direction of the minimum eigenvector of the current information
matrix.  This direction corresponds to the longest axis of the confidence 
ellipsoid defined by the inverse information and is closely related
to E-optimal experimental designs \citep[][]{dette1993geometry}.

In order to balance exploration aimed at reducing the uncertainty about $B^*$ and directly searching for the optimal arm in the initial phase, we use a mixture of information gain criteria, which we refer to as the bound-action mixture (BAM) criterion: 
\begin{equation*}
I_t^{\textnormal{BAM}(\zeta_t(\delta))}(a) = \alpha I_t^{{B}}(a) + (1-\alpha) I_t^{\textnormal{EB-UCB}(\zeta_t(\delta))}(a),
\end{equation*}
where $\alpha \in (0,1)$ is a parameter chosen by the user. Note that while we use the $I_t^{\textnormal{EB-UCB}(\zeta_t(\delta))}$ information gain criterion in this instance, we could use any information gain criterion of choice instead. For notational convenience we drop the $\zeta_t(\delta)$ term and write $I_t^{\textnormal{EB-UCB}}$ for $I_t^{\textnormal{EB-UCB}(\zeta_t(\delta))}$ and $I_t^{\textnormal{BAM}}$ for $I_t^{\textnormal{BAM}(\zeta_t(\delta))}$ since we will use $\zeta_t(\delta) = \min \{\delta, 1/t^2\}$ in the remainder of this manuscript.

Given the advantages of deterministic IDS and its strong performance in 
various experimental settings, we focus on this variant of IDS. Hence, we 
always select 
the action which minimizes the information ratio on the set $\A$, 
as given in \eqref{eq:dids}. 
So at each time step $t \in \{1, \ldots ,T_B \}$ of the bound exploration phase we choose the action
\begin{equation*}
    A_t^{\textnormal{BAM}} = \argmin_{a \in \A} \left\{ \Psi_t^{\textnormal{BAM}}(a) := \frac{\widehat{\Delta}^2_{t, \zeta_t(\delta)}(a)}{I_t^{\textnormal{BAM}}(a)} \right \}. 
\end{equation*}
Throughout the second phase, which we refer to as the \textit{bound exploitation phase}, for any $t \geq T_B + 1$ we use
 \begin{equation*}
     \tilde{B}_t = \min \left\{ B, \min_{\tau \leq t} \left\{ \| \widehat{\bo{\theta}}_{\tau}^{\mathrm{wls}} \|_2 + \beta_{\tau, \zeta_{\tau}(\delta)}^{1/2}(\widehat{B}_{\tau}) \lambda_{\min}(\bo{W}_{\tau})^{-1/2} \right\} \right\}
 \end{equation*}
as the upper bound on $B^*$, with $\widehat{B}_t$ defined in \eqref{eq:hat_B_def}. During this phase we drop the bound information gain criterion $I_t^B$ from the mixture and use only the $I_t^{\textnormal{EB-UCB}}$ criterion.
The quantity $\tilde{B}_t$ is used as the upper bound for $B^*$ for both the gap estimate $\widehat{\Delta}_{t, \zeta_t(\delta)}$ and $I_t^{\textnormal{EB-UCB}}$, which are defined in the same way as in equations \eqref{eq:def_EB_UCB}, \eqref{eq:def_Delta_hat}, and \eqref{eq:def_info_EB_UCB} with $\tilde{B}_t$ in place of $\widehat{B}_t$.
 We summarize this method in Algorithm \ref{algo:ebids}. Note that in the second phase we could use any algorithm which requires explicit use of an upper bound on $B^*$ by taking $B = \tilde{B}_{t}$ as that upper bound. Furthermore, we formulate this procedure specifically in the context of IDS, however, the approach of estimating a high-probability upper bound on the true parameter norm and using it to guide decision making can be thought of as a more general technique, rather than something specific only to IDS.

\begin{algorithm}[ht]\caption{EBIDS}\label{algo:ebids}
   \setstretch{1.0}
Input: Action set $\mathcal{A}$, penalty parameter $\gamma>0$, noise function $\rho: \mathcal{A} \rightarrow \mathbb{R}_{+}$, feature function $\bo{\phi}: \A \rightarrow \mathbb{R^d}$, conservative true parameter norm bound $B$, number of bound exploration steps $T_B$, information gain mixture parameter $\alpha \in (0,1)$, error tolerance parameter $\delta \in (0,1)$.
~~~~
\vspace{2pt}
\begin{algorithmic}[1]
\item[]
\begin{itemize}
\setlength{\itemindent}{-27pt}
    \item[] \setlength{\itemindent}{-27pt}  For $t = 1,2, \ldots, T_B:$
    \begin{itemize}
    \setlength{\itemindent}{-27pt}  
    \item[] Compute $\bo{W}_t$ and $\widehat{\bo{\theta}}^{\mathrm{wls}}_t$ using \eqref{eq:wls_def}
        \item[] $\widehat{B}_t \leftarrow \min \left\{ B, \| \widehat{\bo{\theta}}_{t}^{\mathrm{wls}} \|_2 + \beta_{t, \zeta_t(\delta)}^{1/2}(B) \lambda_{\min}(\bo{W}_{t})^{-1/2} \right\}$
        \item[] $A_t^{\textnormal{EB-UCB}} \leftarrow \argmax_{a \in \A} \left\langle \bo{\phi}(a), \widehat{\bo{\theta}}_t^{\mathrm{wls}}\right\rangle+\beta_{t, \zeta_t(\delta)}^{1 / 2}(\widehat{B}_t)\|\bo{\phi}(a)\|_{\bo{W}_t^{-1}}$
        \item[]  $I_t^{\text{EB-UCB}}(a) \leftarrow \frac{1}{2}\log \left( {\left\| \bo{\phi} (A_t^{\text{EB-UCB}}) \right\|_{\bo{W}_t^{-1}}^2}\right) - \frac{1}{2}\log \left({\left\| \bo{\phi} (A_t^{\text{EB-UCB}}) \right\|_{\left(\bo{W}_t + \rho(a)^{-2} \bo{\phi}(a) \bo{\phi}(a)^{\top}\right)^{-1}}^2}\right)$
        \item[] $I_t^B(a) \leftarrow \frac{1}{2}\log \left( \| \bo{v}_t^{\min} \|_{(\bo{W}_t + \rho(a)^{-2} \bo{\phi}(a) \bo{\phi}(a)^{\top})}^2\right) - \frac{1}{2}\log \left\{ \lambda_{\min} (\bo{W}_t)\right\}$
        \item[] $I_t^{\text{BAM}}(a) \leftarrow \alpha I_t^{{B}}(a) + (1-\alpha) I_t^{\text{EB-UCB}}(a)$
         \item[] $\widehat{\Delta}_{t, \zeta_t(\delta)}(a) \leftarrow \left\langle \bo{\phi}(A_t^{\tn{EB-UCB}}) - \bo{\phi}(a), \widehat{\bo{\theta}}_t^{\mathrm{wls}}\right\rangle+\beta_{t, \zeta_t(\delta)}^{1 / 2}(\widehat{B}_t)\left(\|\bo{\phi}(A_t^{\tn{EB-UCB}})\|_{\bo{W}_t^{-1}} + \| \bo{\phi}(a)\|_{\bo{W}_t^{-1}} \right)$
        \item[] $A_t \leftarrow \argmin_{a \in \A} {\widehat{\Delta}^2_{t, \zeta_t(\delta)}\left(a\right)}/{I_t^{\text{BAM}}\left(a\right)}$
        \item[] Play $A_t$, observe $Y_t  = \left\langle \bo{\phi}(A_t), \bo{\theta}^*\right\rangle+\eta_t$
        \end{itemize}
    \end{itemize}
    \vspace{5pt}
    \begin{itemize}
    \setlength{\itemindent}{-27pt}
\item[] \setlength{\itemindent}{-27pt} For $t = T_B + 1,T_B + 2, \ldots, T:$
    \begin{itemize}
    \setlength{\itemindent}{-27pt}
    \item[] Compute $\bo{W}_t$ and $\widehat{\bo{\theta}}^{\mathrm{wls}}_t$ using \eqref{eq:wls_def}
            \item[] $\widehat{B}_t \leftarrow \min \left\{ B, \| \widehat{\bo{\theta}}_{t}^{\mathrm{wls}} \|_2 + \beta_{t, \zeta_t(\delta)}^{1/2}(B) \lambda_{\min}(\bo{W}_{t})^{-1/2} \right\}$
            \item[] $\tilde{B}_t \leftarrow \min \left\{ B, \min_{\tau \leq t} \left\{ \| \widehat{\bo{\theta}}_{\tau}^{\mathrm{wls}} \|_2 + {\beta}_{\tau, \zeta_{\tau}(\delta)}^{1/2}(\widehat{B}_t) \lambda_{\min}(\bo{W}_{\tau})^{-1/2} \right\} \right\}$
        \item[] $A_t^{\textnormal{EB-UCB}} \leftarrow \argmax_{a \in \A} \left\langle \bo{\phi}(a), \widehat{\bo{\theta}}_t^{\mathrm{wls}}\right\rangle+\beta_{t, \zeta_t(\delta)}(\tilde{B}_t)^{1 / 2}\|\bo{\phi}(a)\|_{\bo{W}_t^{-1}}$
        \item[]  $I_t^{\text{EB-UCB}}(a) \leftarrow \frac{1}{2}\log \left( {\left\| \bo{\phi} (A_t^{\text{EB-UCB}}) \right\|_{\bo{W}_t^{-1}}^2}\right) - \frac{1}{2}\log \left({\left\| \bo{\phi} (A_t^{\text{EB-UCB}}) \right\|_{\left(\bo{W}_t + \rho(a)^{-2} \bo{\phi}(a) \bo{\phi}(a)^{\top}\right)^{-1}}^2}\right)$
        \item[] $\widehat{\Delta}_{t, \zeta_t(\delta)}(a) \leftarrow \left\langle \bo{\phi}(A_t^{\tn{EB-UCB}}) - \bo{\phi}(a), \widehat{\bo{\theta}}_t^{\mathrm{wls}}\right\rangle+\beta_{t, \zeta_t(\delta)}^{1/2}(\tilde{B}_t)\left(\|\bo{\phi}(A_t^{\tn{EB-UCB}})\|_{\bo{W}_t^{-1}} + \| \bo{\phi}(a)\|_{\bo{W}_t^{-1}} \right)$
        \item[] $A_t \leftarrow \argmin_{a \in \A} {\widehat{\Delta}^2_{t, \zeta_t(\delta)}\left(a\right)}/{I_t^{\text{EB-UCB}}\left(a\right)}$
        \item[] Play $A_t$, observe $A_t = \left\langle \bo{\phi}(A_t), \bo{\theta}^*\right\rangle+\eta_t$
        \end{itemize}
    \end{itemize}
    \end{algorithmic}
    \end{algorithm}

\section{Regret analysis of EBIDS algorithm}\label{regret_bound}
In this section we present the regret and pseudo-regret bounds for both phases of the EBIDS algorithm. We defer the proofs of these propositions and relevant lemmas to the Supplementary Materials.
For any $t$ and $\xi_t > 0$, let $E_{t, \xi_t}$ be the event
\begin{equation}\label{eq:event_E_t}  E_{t, \xi_t} = \left\{\left\| \bo{\theta}^* - \widehat{\bo{\theta}}_t^{\mathrm{wls}} \right \|_{\bo{W}_t}^2 \leq \beta_{t, \xi_t}(B^*)\right\},
\end{equation}
and define $E_{\delta} = \bigcap^{\infty}_{t=1} E_{t, \delta}$. Note that by Theorem \ref{thm:snmb} we have $\Prob(E_{\delta}) \geq 1-\delta$.  The following proposition summarizes the regret and pseudo-regret bounds for EBIDS during the bound exploration phase. 
\begin{prop}\label{prop:regret_1}
    For any $2 \leq T \leq T_B$ the regret $\mathcal{R}_T$ of Algorithm \ref{algo:ebids} is bounded above by 
    \begin{equation*}
            \mathcal{R}_T  \leq O\left(\frac{d \max \{U / \sqrt{\gamma}, \rho_{\max}\}}{\sqrt{1-\alpha}} \sqrt{T} \log T \sqrt{ \log (1/\delta) + \log\left(1 + \frac{\ru}{\gamma}\right) + \gamma B^2 }  \right)
    \end{equation*}
    and whenever event $E_{\delta}$ holds the pseudo-regret $ \mathcal{PR}_T$ is bounded above by the same rate.
\end{prop}

We also provide guarantees on the estimated upper bound on $B^*$ after the bound exploration phase. This, in turn, will allow us to obtain an improved bound for the regret and pseudo-regret in the subsequent phase.

\begin{prop}\label{prop:high_prob_B}
For any constant $g > 0$, with sufficiently large $T_B$ and sufficiently large $\alpha$, whenever event $E_{\delta}$ holds we have $B^* \leq \tilde{B}_t \leq (1+g)B^*$
for any $t \geq T_B+1$.
\end{prop}
Please see Section \ref{proof_high_prob_B} in the Supplementary Materials for the exact constants required as lower bounds for $T_B$ and $\alpha$ depending on $g$. 
Finally, using the results of Proposition \ref{prop:high_prob_B} we are able to establish a regret bound for the second phase of EBIDS which is independent of the original conservative bound $B$.
\begin{prop}\label{prop:regret_2}
For any constant $g > 0$, with sufficiently large $T_B$ and sufficiently large $\alpha$, with probability at least $1-\delta$ the regret and pseudo-regret of Algorithm \ref{algo:ebids} are both bounded above by $O\left(d U \rho_{\max}(1+g)B^* \sqrt{T} \log T\right)$, for any $T \geq T_B+1$. 
\end{prop}
Similarly, we give the exact constants required as lower bounds for $T_B$ and $\alpha$ in Supplementary Materials, in Section \ref{proof_regret_2}. Thus, Propositions \ref{prop:regret_1} and \ref{prop:regret_2} together give us regret and pseudo-regret guarantees for both bound exploration phase and the subsequent bound exploitation phase of EBIDS. This is different from \citet{norm_agnostic} who do not control the regret in the initial stages of their norm-agnostic algorithms.

\section{Simulation study}\label{simulation_study}
We evaluate the performance of EBIDS using simulation studies and compare it with the norm-agnostic competitor algorithms NAOFUL and OLSOFUL by \citet{norm_agnostic} which also aim at alleviating the dependence on access to a high-quality bound on the true parameter norm. We include the EB-UCB algorithm to demonstrate the advantage of using the IDS strategy in addition to utilizing the empirical norm bound. We run the comparison also against the {oracle} versions of EBIDS, IDS-UCB and UCB with access to the true value of $B^*$. We use the same setting as in the simulation illustration in Section \ref{information_directed_sampling} with 
$\bo{\theta}^* = [-5, 1, 1, 1.5, 2]^{\top}$ as the true parameter and ten arms with features sampled from Unif$[-1/\sqrt{5}, 1/\sqrt{5}]$.
The error distribution for the first five arms
are standard normal and for the remaining five arms they are normal with mean zero and standard deviation $0.2$. We take the conservative $B=100$ as the assumed upper bound on $B^*$. Both the oracle and non-oracle versions of EBIDS use $\alpha=0.5$, giving equal weight to both components of the BAM criterion, and run the bound exploration phase for $T_B = 50$ steps.

\begin{figure}[ht]
    \begin{center}
        \includegraphics[scale=0.38]{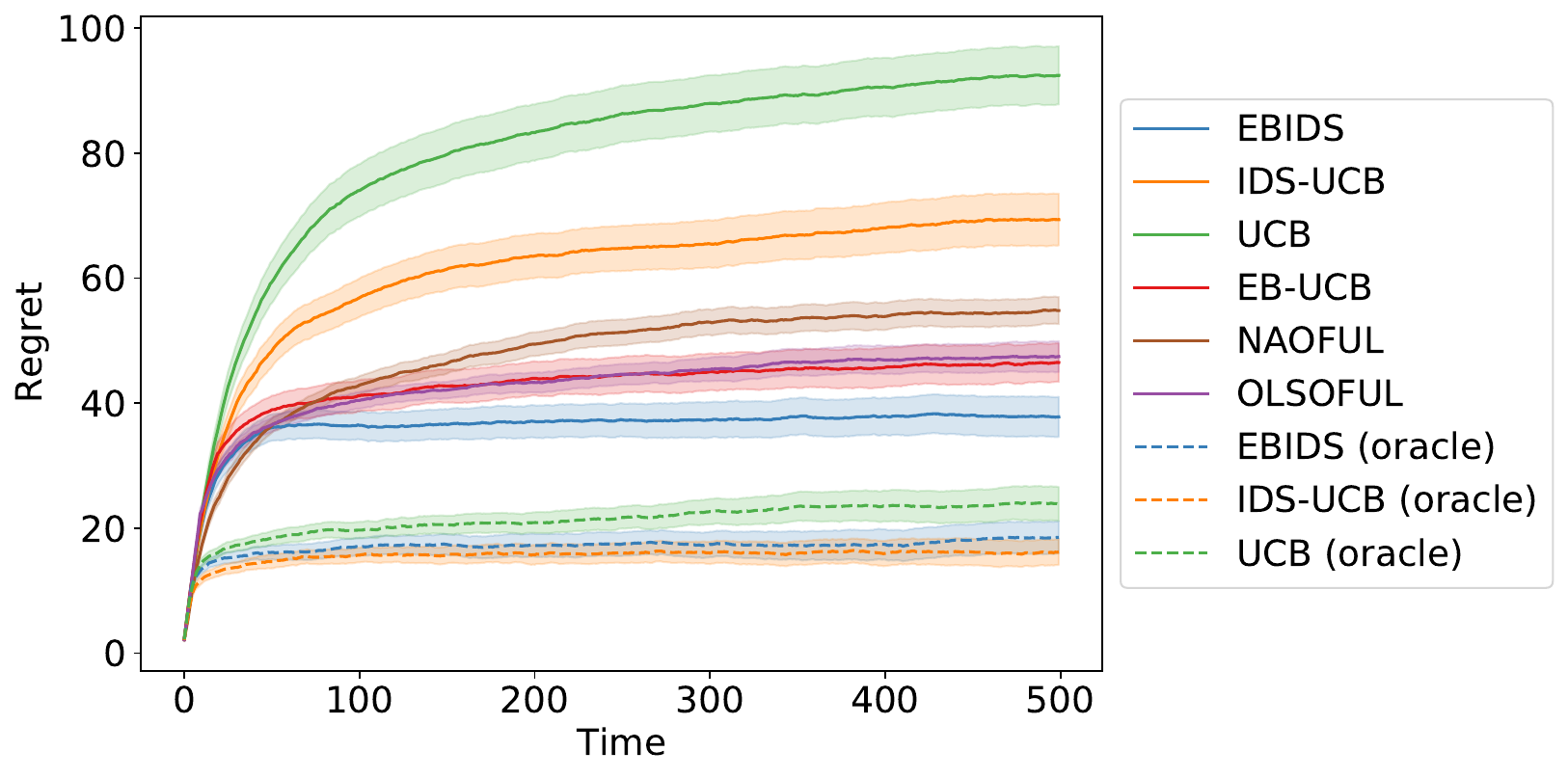}
    \end{center}
    \caption{Regret incurred by EBIDS, EB-UCB, NAOFUL, OLSOFUL, IDS-UCB and UCB with conservative $B=100$. We include the oracle versions of EBIDS, IDS-UCB, and UCB using $B=B^*$ for reference. The solid and dashes lines represent the regret averaged over $200$ repeated experiments, while the shaded bounds represent $95\%$ pointwise confidence bounds.}
    \label{fig:algos_comparison}
\end{figure}

Figure \ref{fig:algos_comparison} shows the mean regret averaged over $200$ repeated experiments with $T=500$ steps
along with $95\%$ pointwise confidence bounds. 
As we can see, EB-UCB is competitive with NAOFUL and OLSOFUL, while EBIDS performs best among all the algorithms which do not have access to the true parameter norm. It achieves significantly lower regret than IDS-UCB and UCB. Meanwhile, the performance of oracle EBIDS is better than that of oracle UCB and almost indistinguishable from the one achieved by oracle IDS-UCB.

\begin{figure}[ht]
    \begin{center}
        \includegraphics[scale=0.38]{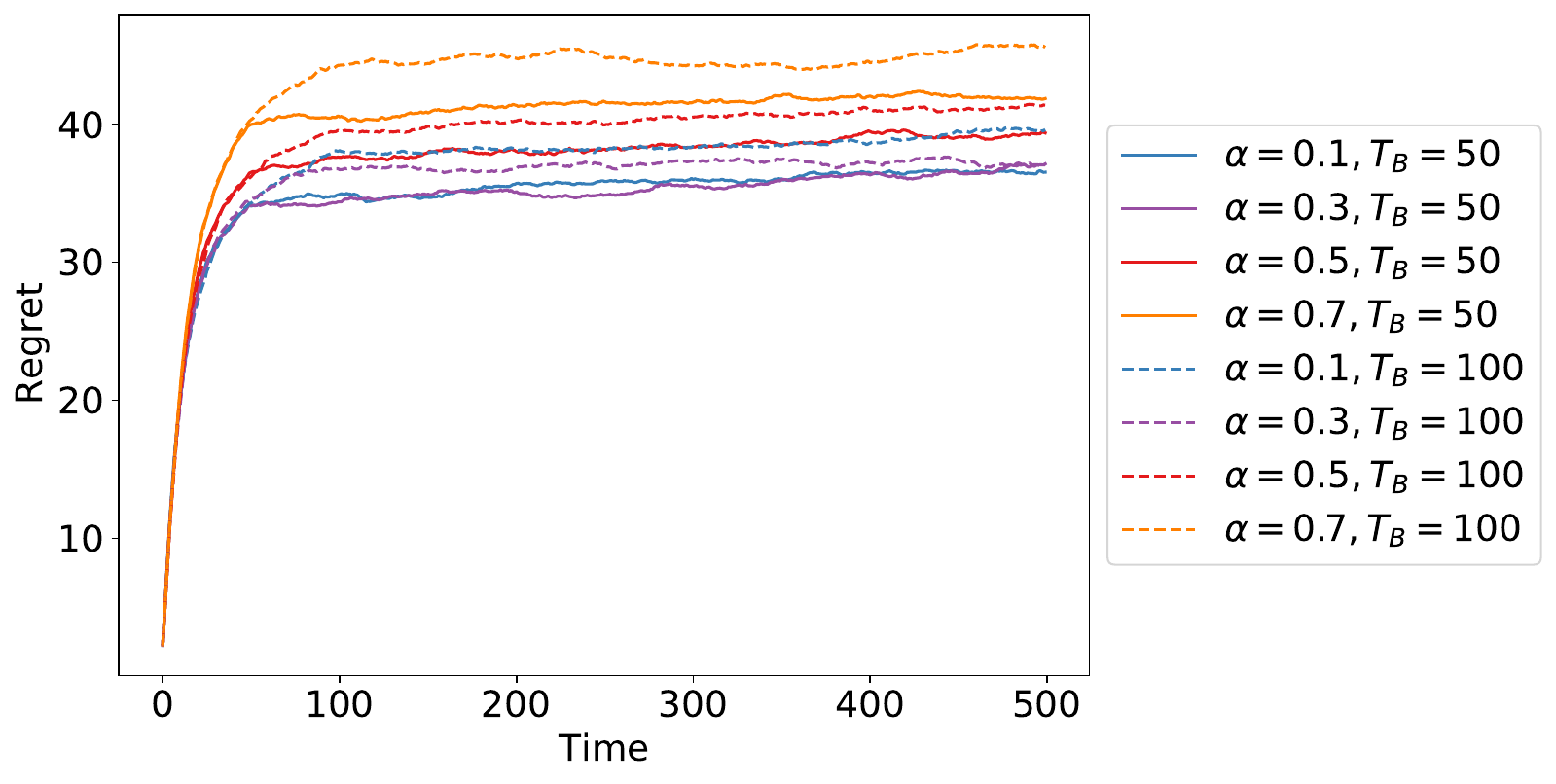}
    \end{center}
    \caption{Average regret for EBIDS averaged over $200$ repeated experiments with $T=500$ steps under different values of the tuning parameter $\alpha$ and the length $T_B$ of the bound exploration phase.}
    \label{fig:ablation}
\end{figure}

We also perform an ablation study to determine the sensitivity of EBIDS to the tuning parameter $\alpha$ and the length $T_B$ of the bound exploration phase. We consider all combinations of $\alpha \in \{0.1, 0.3, 0.5, 0.7\}$ and $T_B \in \{50, 100\}$. We use the same setting as above and present the results for $T=500$ steps averaged over $200$ repeated experiments in Figure \ref{fig:ablation}. Using $T_B = 50$ leads to somewhat better results than $T_B=100$ and $\alpha = 0.3$ performs best for both values of $T_B$. However, the performance is similar for all considered combinations of the tuning parameters, especially compared to the differences in performance of the competitor algorithms. 
This shows that while EBIDS, like most other bandit algorithms, uses tuning parameters, its performance is not very sensitive to their choice, with several considered combinations of $\alpha$ and $T_B$ achieving practically indistinguishable regret.

\section{Discussion}

Bandit algorithms often require access to a high-quality upper bound on the Euclidean norm of the true parameter vector in order to achieve good performance. In practice, such information is rarely available 
{\em a priori}, which can lead to significant regret accumulation. Despite its prevalence, this problem has received relatively little attention in the bandit literature. We introduced the empirical bound information-directed sampling (EBIDS) algorithm which addresses this challenge by iteratively refining a high-probability upper
bound on the true parameter norm. We developed a novel information criterion that balances 
tightening the bound on the true parameter norm and explicitly searching for the optimal arm. In
simulation experiments, EBIDS showed improved performance compared to the competing norm-agnostic algorithms. Furthermore, we proved regret bounds that eventually do not depend on the initially assumed bound for the parameter norm, and unlike prior regret guarantees, our bounds are anytime in that they apply 
to all phases of the algorithm. 


\subsubsection*{Broader Impact Statement}
This paper introduces novel methodology for frequentist IDS that does not 
require strong prior information on the norm of the true parameter indexing
the reward model.  Our methodology, which involves a novel information criterion,
can be viewed as a general approach to balancing bound improvement 
and regret minimization that is applicable in a wide
range of UCB and IDS bandit algorithms.



\bibliography{main}
\bibliographystyle{rlj}

\beginSupplementaryMaterials

In these supplementary materials we provide the proofs to the propositions we have stated in the paper as well as some additional simulation studies results.

\section{Notation and lemmas}\label{notation_and_lemmas}
We begin by introducing some notation and basic facts.
For any unit vector $\bo{v} \in \R^d$ and any $a \in \A$, let ${\psi}_{\bo{v}}(\bo{\phi}(a)), {\psi}^{\perp}_{\bo{v}}(\bo{\phi}(a)) \in \R$ denote the orthogonal decomposition of $\bo{\phi}(a)$, i.e.,  
$$\bo{\phi}(a) = {\psi}_{\bo{v}}(\bo{\phi}(a))\bo{v} + {\psi}^{\perp}_{\bo{v}}(\bo{\phi}(a))\bo{v}^{\perp},$$ 
where $\| \bo{v}^{\perp}\|_2 = 1$ and $\bo{v}^{\perp} \perp \bo{v}$.
Let 
\begin{equation}\label{eq:kappa_def}
\kappa = \min_{\bo{v} \in \R^d \text{ s.t. } \| \bo{v} \|_2 = 1} \max_{a \in \A} \left\{\rho(a)^{-2} {\psi}_{\bo{v}}(\bo{\phi}(a))^2 \right \}.
\end{equation}
Note that $\kappa > 0$.
Let
\begin{equation}\label{eq:omega_def}
\omega_t(a) = \rho(a)^{-2}\psi_{\bo{v}_t^{\min}}(\phi(a))^2 . 
\end{equation}

Also, note that for any $a \in \A$ we have 
\begin{equation*}
\| \bo{\phi}(a)\|_{\bo{W}_t^{-1}}^2 = \sum_{i = 1}^d \psi_{\bo{v}_i}(\bo{\phi}(a))^2 \lambda_i^{-1}
\end{equation*}
where $\{(\lambda_i, \bo{v}_i)\}_{i=1}^d$ are the eigenvalue-eigenvector pairs of $\bo{W}_t$. 
Hence for every $t \geq 1$ and $a \in \A$ we have
\begin{equation*}
    \| \bo{\phi}(a) \|_2^2 \lambda_{\max}(\bo{W}_t)^{-1} \leq \| \bo{\phi}(a) \|_{\bo{W}_t^{-1}}^2 \leq \| \bo{\phi}(a) \|_2^2 \lambda_{\min}(\bo{W}_t)^{-1},
\end{equation*}
so
\begin{equation}\label{eq:11-basic}
    L^2 \lambda_{\max}(\bo{W}_t)^{-1} \leq \| \bo{\phi}(a) \|_{\bo{W}_t^{-1}}^2 \leq U^2 \lambda_{\min}(\bo{W}_t)^{-1}.
\end{equation}

Also from Cauchy-Schwarz inequality 

\begin{equation}\label{eq:star_cauchy_schwarz}
    \left\langle \bo{\phi}(a), \widehat{\bo{\theta}}_t^{\mathrm{wls}} \right\rangle^2 \leq \left\| \bo{\phi}(a)\right\|_2^2 \left\| \widehat{\bo{\theta}}_t^{\mathrm{wls}}\right\|_2^2 \leq U^2 \left \| \widehat{\bo{\theta}}_t^{\mathrm{wls}} \right\|_2^2
\end{equation}

From Weyl's inequality \citep{matrix_theory}, for any positive semi-definite matrices $\bo{A}, \bo{B}$ we have 
\begin{equation*}
\lambda_{\max}(\bo{A} + \bo{B}) \leq \lambda_{\max}(\bo{A}) + \lambda_{\max}(\bo{B}). 
\end{equation*}
Thus, for every $t \geq 1$ we have
\begin{equation}\label{eq:max_eig_bound}
    \lambda_{\max}(\bo{W}_t) \leq \lambda_{\max}(\gamma \bo{I}_d) + \sum_{\tau=1}^{t-1} \lambda_{\max}(\rho(a_{\tau})^{-2} \bo{\phi}(a_{\tau}) \bo{\phi}(a_{\tau})^{\top}) \leq \gamma + (t-1)\rho_{\min}^{-2}U^2,
\end{equation}
so from \eqref{eq:11-basic}, for any $t \geq 1$ we have
\begin{equation}\label{eq:13-basic}
    \| \bo{\phi}(a) \|_{\bo{W}_t^{-1}}^2 \geq \frac{L^2}{\gamma + (t-1)\ru} \geq \frac{L^2}{t(\gamma + \ru)}.
\end{equation}

Also from \eqref{eq:max_eig_bound} for $T \geq 2$ we have
\begin{align}\label{eq:bound_log_det_frac}
    \log\left(\frac{\det (\bo{W}_T)}{\det (\bo{W}_1)}\right) = & \log (\det (\bo{W}_T)) - \log (\det( \gamma \bo{I}_d)) \leq d  \log (\gamma + (T-1)\ru) - d \log \gamma \nonumber \\
    = & d \log \left[1 + (T-1)\frac{\ru}{\gamma}\right] 
    \leq d \log \left[(T-1)\left(1+ \frac{\ru}{\gamma}\right)\right] \nonumber \\
    = & d \log(T-1) + d \log\left(1 + \frac{\ru}{\gamma}\right).
\end{align}
Applying the data processing inequality \citep{cover2012elements} in an analogous way as \citet{kirschner2018}, we obtain
\begin{align}\label{eq:eb_ucb_info_bound}
    I_t^{\tn{EB-UCB}}(a) \leq \frac{1}{2}\log\left( \frac{\det(\bo{W}_t + \rho(a)^{-2} \bo{\phi}(a) \bo{\phi}(a)^{\top})}{ \det(\bo{W}_{t})}\right) = \frac{1}{2}\log \left(1 + \rho(a)^{-2} \| \bo{\phi}(a) \|^2_{\bo{W}_t^{-1}}\right)
\end{align}
for any $a \in \A$. 
So from \eqref{eq:bound_log_det_frac} we get
\begin{align}\label{eq:bound_UCB_info}
    \sum_{t=1}^T I_t^{\tn{EB-UCB}}(a_t) \leq \frac{1}{2}\log\left(\frac{\det (\bo{W}_{T+1})}{\det (\bo{W}_1)}\right) \leq \frac{1}{2} d \log T + \frac{1}{2} d\log\left(1 + \frac{\ru}{\gamma}\right) = O(d \log T),
\end{align}
for any sequence $\{a_t\}_{t=1}^{T} \subset \A$.

We now state and prove some additional lemmas that will be useful throughout the proofs of Propositions \ref{prop:regret_1} - \ref{prop:regret_2}.

\begin{lemma}\label{lem:ids_mixtures}
  Let $\widehat{\Delta}_t: \A \rightarrow \R_{+}$ be a gap estimate function and let $I_t^X, I_t^Y : \A \rightarrow \R_{+}$ be two information gain criteria. Let $I_t^{XY}$ be the mixture information gain criterion given by 
  $$I_t^{XY}(a) = \alpha I_t^X(a) + (1 - \alpha) I_t^Y(a)$$
  for some $\alpha \in (0,1)$. Consider now a deterministic IDS algorithm which at each time step $t$ plays action $a_t^{XY}$ given by
  $$
  a_t^{XY} = \argmin_{a \in \A} \frac{\widehat{\Delta}^2_{t}(a)}{I_t^{XY}(a)}
  $$
  Then at each time step $t$ the information gain on to the first criterion $I_t^X$ is lower-bounded by
  $$
  I_t^X\left(a_t^{XY}\right) \geq \frac{\widehat{\Delta}^2_{t}\left(a^{XY}_t\right)}{\widehat{\Delta}^2_{t}\left(a^{{I}, X}_t\right)} I_t^X\left(a_t^{{I}, X}\right) - \frac{1-\alpha}{\alpha}I_t^Y\left( a_t^{XY}\right),
  $$
  where
  $a_t^{I, X} = \argmax_{a \in \A} I_t^X(a)$.
  
\end{lemma}

\begin{lemma}\label{lem:eigvals}
  Recall the definition $\omega_t(a) = \rho(a)^{-2}{\psi}_{\bo{v}_t^{\min}}(\bo{\phi}(a))^2 $. For any $T \geq 1$ and any sequence of actions $\{a_t\}_{t=1}^{T}$ we have
  $$\lambda_{\min}(\bo{W}_{T+1}) \geq \gamma - \ru + \frac{1}{d}\sum_{t=1}^T \omega_t(a_t).$$
\end{lemma}

\begin{lemma}\label{lem:sequences}
Let $\{x_t\}_{t=1}^{T+1} \subset [0, U]$ be a bounded sequence  for some constant $U > 0$. Then for any constant $c > 0$ we have 
$$\sum_{t=1}^T \frac{x_{t+1}}{c + \sum_{\tau = 1}^{t} x_{\tau}} \leq \log T + \frac{U}{c} + 1.$$ 
\end{lemma}

\section{Proofs of theoretical results}

In this section we provide the proofs of Theorem \ref{thm:univ_IDS_bound}, Lemmas \ref{lem:ids_mixtures} - \ref{lem:sequences}, and Propositions \ref{prop:regret_1} - \ref{prop:regret_2}.
\subsection{Proof of Theorem \ref{thm:univ_IDS_bound}} \label{proof_thm_1}
Recall that by Cauchy-Schwarz inequality, for any random variables $\{X_t\}_{t \in G}, \{ Y_t\}_{t \in G}$ with nonnegative support, with probability $1$ we have 
\begin{equation*}
\sum_{t \in G}\sqrt{X_tY_t} \leq \sqrt{\left(\sum_{t \in G}X_t\right)\left(\sum_{t \in G} Y_t\right)},
\end{equation*}
and for any random variables $X, Y$ with nonnegative support we have 
\begin{equation*}
    \E \left[\sqrt{XY}\right] \leq \sqrt{\E [X] \E[Y]}.
\end{equation*}
Hence if $\widehat{\Delta}(A_t) \geq \Delta(A_t)$, for all $t \in G$, then with probability $1$ we have
$$\sum_{t \in G} {\Delta}\left(A_t\right) \leq \sum_{t \in G} \widehat{\Delta}_t\left(A_t\right) = \sum_{t \in G}\sqrt{\Psi_t(A_t) I_t(A_t)} \leq \sqrt{\left[\sum_{t \in G}\Psi_t(A_t)\right] \left[\sum_{t \in G} I_t(A_t)\right]}.$$
Also
\begin{align*}
\E\left[\sum_{t \in G} \widehat{\Delta}_t\left(A_t\right) \right] = &
 \E \left[\sum_{t \in G} \sqrt{\Psi_t(A_t) I_t\left(A_t\right)}\right] \leq
 \E \left(\sqrt{\left[\sum_{t \in G} \Psi_t\left(A_t\right)\right]\left[\sum_{t \in G} I_t\left(A_t\right)\right]} \right) \\
 \leq & \sqrt{\E \left[\sum_{t \in G} \Psi_t\left(A_t\right)\right]\E \left[\sum_{t \in G} I_t\left(A_t\right)\right] }. \qed
\end{align*}


\subsection{Proof of Lemma \ref{lem:ids_mixtures}}
By the definition of $a_t^{XY}$ we have 
$$\frac{\widehat{\Delta}_t^2(a_t^{XY})}{\alpha I_t^X(a_t^{XY}) + (1 - \alpha) I_t^Y(a_t^{XY})} \leq \frac{\widehat{\Delta}_t^2(a_t^{{I}, X})}{\alpha I_t^X(a_t^{{I}, X}) + (1 - \alpha) I_t^Y(a_t^{{I}, X})}, $$
hence
$$
\alpha I_t^X(a_t^{XY}) + (1 - \alpha) I_t^Y(a_t^{XY}) \geq \frac{\widehat{\Delta}_t^2(a_t^{XY})}{\widehat{\Delta}_t^2(a_t^{\textnormal{I-}X})}\left[ \alpha I_t^X(a_t^{{I}, X}) + (1-\alpha)I_t^Y (a_t^{{I}, X}) \right],
$$
and thus
\begin{align*}
I_t^X(a_t^{XY}) \geq & \frac{\widehat{\Delta}_t^2(a_t^{XY})}{\widehat{\Delta}_t^2(a_t^{{I}, X})}I_t^X(a_t^{{I}, X}) + \frac{(1-\alpha)}{\alpha}\cdot \frac{\widehat{\Delta}_t^2(a_t^{XY})}{\widehat{\Delta}_t^2(a_t^{{I}, X})}I_t^Y (a_t^{{I}, X}) - \frac{1-\alpha}{\alpha}  I_t^Y(a_t^{XY}) \\
\geq & \frac{\widehat{\Delta}_t^2(a_t^{XY})}{\widehat{\Delta}_t^2(a_t^{{I}, X})}I_t^X(a_t^{{I}, X}) - \frac{1-\alpha}{\alpha}  I_t^Y(a_t^{XY}). \qed
\end{align*}

\subsection{Proof of Lemma \ref{lem:eigvals}}

Recall that we define $\lambda_{1}^{(t)}, \ldots , \lambda_{d}^{(t)}$ as the (not necessarily ordered) eigenvalues of $\bo{W}_t$. Let 
$$i^*(t) = \argmin_{1\leq i \leq d} \lambda_i^{(t)}.$$
By Weyl's inequality \citep{matrix_theory}, for any symmetric positive semi-definite matrices $\bo{A}, \bo{B} \in \R^{m \times m}$ we have 
\begin{equation}\label{eq:eigs_sum_matrices}
    \lambda_{(i)}(\bo{A} + \bo{B}) \geq \lambda_{(i)}(\bo{A})
\end{equation}
where $\lambda_{(i)}(\bo{A})$ is the $i$-th largest eigenvalue of $\bo{A}$ for any $1 \leq i \leq m$. Let $\bo{v}_t^{\min}$ be the unit eigenvector corresponding to the smallest eigenvalue of $\bo{W}_t$. Then for any $1\leq i \leq d$ we have 
\begin{align*}
    \lambda_{(i)}(\bo{W}_{t+1}) =& \lambda_{(i)}\left(\bo{W}_t +  \rho(a_t)^{-2}\bo{\phi}(a_t)\bo{\phi}(a_t)^{\top} \right) \\
    =& \lambda_{(i)}\left(\bo{W}_t + \rho(a_t)^{-2}{\psi}_{\bo{v}_t^{\min}}(\bo{\phi}(a))\bo{v}_t^{\min}(\bo{v}_t^{\min})^{\top} \right. \\
    &+ \left. \rho(a_t)^{-2}{\psi}^{\perp}_{\bo{v}_t^{\min \perp}}(\bo{\phi}(a))\bo{v}_t^{\min \perp}(\bo{v}_t^{\min \perp})^{\top}\right) \\
    \geq & \lambda_{(i)}\left(\bo{W}_t + \rho(a_t)^{-2}{\psi}_{\bo{v}_t^{\min}}(\bo{\phi}(a))\bo{v}_t^{\min}(\bo{v}_t^{\min})^{\top} \right) \\
= & \lambda_{(i)}\left(\bo{W}_t + \omega_t(a_t)\bo{v}_t^{\min}(\bo{v}_t^{\min})^{\top} \right).
\end{align*}
Note that the matrix $\bo{W}_t + \omega_t(a_t)\bo{v}_t^{\min}(\bo{v}_t^{\min})^{\top}$ has the same eigenvectors as $\bo{W}_t$ and the smallest eigenvalue of $\bo{W}_t$, i.e., the one corresponding to $\bo{v}_t^{\min}$ is increased by $\omega_t(a_t)$. So for any $t$ we can order the eigenvalues $\lambda_1^{(t+1)}, \ldots , \lambda_d^{(t+1)}$ in such way that $\lambda_i^{(t+1)} \geq \lambda_i^{(t)}$ and 
$$
\lambda_{i^*(t)}^{(t+1)} \geq \lambda_{i^*(t)}^{(t)} + \omega_t(a_t). 
$$
Since we have $d$ eigenvalues and at each time step $t$ we add at least $\omega_t(a_t)$ to the smallest eigenvalue at that time step without reducing the other ones we have
$$
\lambda_{i^*(T)}^{(T)} - \lambda_{i^*(1)}^{(1)} + \omega_T(a_T) \geq \frac{1}{d} \sum_{t=1}^T \omega_t(a_t).
$$
Note that $\lambda_1^{(1)} = \ldots = \lambda_d^{(1)} = \gamma$ and 
$\omega_T(a_T) \leq \ru$, so 

$$\lambda_{\min}(\bo{W}_{T+1})  = \lambda_{i^*(T+1)}^{(T+1)} \geq \lambda_{i^*(T)}^{(T)} \geq  \gamma - \ru + \frac{1}{d}\sum_{t=1}^T \omega_t(a_t). \qed$$



\subsection{Proof of Lemma \ref{lem:sequences}}

Let 
$$f(x_1, \ldots, x_{T+1}) = \sum_{t=1}^T \frac{x_{t+1}}{c + \sum_{\tau = 1}^{t} x_{\tau}}$$ 
We use induction to show that the maximum of $f$ is achieved at $x_1 = 0$ and $$x_2 = x_3 = \ldots = x_{T+1} = U.$$ Note that for any $\tilde{x}_1, \ldots , \tilde{x}_T \in [0,U]$ we have 
$$\argmax_{{x}_{T+1} \in [0,U]} f(\tilde{x}_1, \ldots , \tilde{x}_{T}, x_{T+1}) = U.$$ 
Suppose that for any $t \geq 2$ it holds that for any $t \leq k \leq T$ and any $\tilde{x}_1, \ldots , \tilde{x}_k \in [0,U]$ we have 
\begin{equation}\label{eq:backwards}
    (x_{k+1}^*, \ldots ,x_{T+1}^*) := \argmax_{{x}_{k+1}, \ldots , {x}_{T+1} \in [0,U]} f(\tilde{x}_1, \ldots, \tilde{x}_k, {x}_{k+1}, \ldots, {x}_{T+1}) = (U, \ldots, U) \in \mathbb{R}^{T-k+1}.
\end{equation}

Take any $\tilde{x}_1, \ldots , \tilde{x}_{t-1} \in [0,U]$. Then by taking $k = t+1$ the above statement gives us
\begin{equation*}
    \max_{x_t, x_{t+1}, \ldots , x_{T+1} \in [0,U]} f(\tilde{x}_1, \ldots , \tilde{x}_{t-1}, x_t, x_{t+1}, \ldots, x_{T+1}) = \max_{x_t, x_{t+1} \in [0, U]} f(\tilde{x}_1, \ldots , \tilde{x}_{t-1}, x_t, x_{t+1}, U, \ldots , U).
\end{equation*}
Let 
\begin{align*}
(\check{x}_t, \check{x}_{t+1}) = \argmax_{x_t, x_{t+1} \in [0,U]} f(\tilde{x}_1, \ldots , \tilde{x}_{t-1}, x_t, x_{t+1}, \ldots, x_{T+1}).
\end{align*}
Note that $\check{x}_{t+1} = x^*_{t+1} = U$ by taking the induction statement with $k = t$. 
For notational convenience let $b = c + \sum_{\tau = 1}^{t-1} \tilde{x}_{\tau}$. Then 
\begin{align*}
(\check{x}_t, \check{x}_{t+1})
= & \argmax_{x_t, x_{t+1} \in [0,U]} \left\{ \frac{x_t}{b} + \frac{x_{t+1}}{b + x_t} + \sum_{\tau=0}^{T-t-1}\frac{U}{b + x_t + x_{t+1} + \tau U}  \right \}.
\end{align*}
Let 
\begin{equation*}
g_t(x_t, x_{t+1}) =  \frac{x_t}{b} + \frac{x_{t+1}}{b + x_t} + \sum_{\tau=0}^{T-t-1}\frac{U}{b + x_t + x_{t+1} + \tau U}.
\end{equation*}
Suppose that $\check{x}_t = x$ for some $0 \leq x < U$. Note that 
\begin{equation*}
    g_t(U,x) - g_t(x, U) = \left(\frac{U}{b} + \frac{x}{b+U}\right) - \left(\frac{x}{b} + \frac{U}{b+x}\right) = \frac{Ux(U-x)}{b(b + U)(b + x)} > 0.
\end{equation*}
So $g_t(U,x) > g_t(x,U) = g_t(\check{x}_t, \check{x}_{t+1})$ which is a contradiction, since $(\check{x}_t, \check{x}_{t+1})$ is the maximizer of $g_t(x_t, x_{t+1})$. So $\check{x}_t = U$. Thus, we have shown that for any $\tilde{x}_1, \ldots , \tilde{x}_{t-1} \in [0,U]$ we have 
$$(x_{t}^*, \ldots ,x_{T+1}^*) = \argmax_{{x}_{t}, \ldots , {x}_{T+1} \in [0,U]} f(\tilde{x}_1, \ldots, \tilde{x}_{t-1}, {x}_{t}, \ldots, {x}_{T+1}) = (U, \ldots, U) \in \mathbb{R}^{T-t+2}.$$
Hence by induction we get that for any $\tilde{x}_1 \in [0,U]$ we have 
$$\argmax_{{x}_{2}, \ldots , {x}_{T+1} \in [0,U]} f(\tilde{x}_1, x_2, \ldots, {x}_{T+1}) = (U, \ldots, U) \in \mathbb{R}^{T}.$$
Clearly
$$
\argmax_{x_1 \in [0,U]} f(x_1, U, \ldots , U) = 0,
$$
so
\begin{align*}
\max_{{x}_{1}, \ldots , {x}_{T+1} \in [0,U]} f({x}_1, \ldots, {x}_{T+1}) =& f(0, U, \ldots , U) = \sum_{t=1}^T \frac{U}{c + (t-1)U} \nonumber \\
\leq & \frac{U}{c} + \sum_{t=2}^{T} \frac{1}{t-1} < \log T + \frac{U}{c} + 1. \qed
\end{align*}

\subsection{Proof of Proposition \ref{prop:regret_1}}


From Theorem \ref{thm:univ_IDS_bound}, for any $T \leq T_B$ we have

\begin{align*}
    \E \left[ \sum_{t=1}^T \widehat{\Delta}_{t,\zeta_t(\delta)} (A_t^{\textnormal{BAM}})\right] \leq &  \sqrt{\left( \E \left[\sum_{t=1}^T \Psi^{\textnormal{BAM}}_t \left(A_t^{\textnormal{BAM}}\right) \right]\right)\left(\E \left[\sum_{t=1}^T I^{\tn{BAM}}_t\left(A_t^{\tn{BAM}}\right)\right]\right)} \\
    \leq & \sqrt{\left( \E \left[\sum_{t=1}^T \frac{\widehat{\Delta}_{t, \zeta_t(\delta)}^2\left(A_t^{\textnormal{BAM}}\right)}{I_t^{\tn{BAM}}\left(A_t^{\textnormal{BAM}}\right)} \right]\right)\left(\E \left[\sum_{t=1}^T I^{\tn{BAM}}_t\left(A_t^{\tn{BAM}}\right)\right]\right)} \\
    =& \sqrt{\E \left[\sum_{t=1}^T \frac{\widehat{\Delta}^2_{t,\zeta_t(\delta)}\left(A_t^{\tn{BAM}}\right)}{\alpha I_t^{B}\left(A_t^{\tn{BAM}}\right) + (1-\alpha) I_t^{\tn{EB-UCB}}\left(A_t^{\tn{BAM}}\right) }\right]} \\
    \times & \sqrt{{\alpha \E \left[ \sum_{t=1}^T I_t^{B}\left(A_t^{\tn{BAM}}\right)\right] + (1-\alpha) \E \left[\sum_{t=1}^T I_t^{\tn{EB-UCB}}\left(A_t^{\tn{BAM}}\right)\right] }}.
\end{align*}
By the definition of $A_t^{\tn{BAM}}$ we have
\begin{align}\label{eq:Psi_exp_1}
    \E \left[\sum_{t=1}^T \frac{\widehat{\Delta}^2_{t,\zeta_t(\delta)}\left(A_t^{\tn{BAM}}\right)}{\alpha I_t^{B}\left(A_t^{\tn{BAM}}\right) + (1-\alpha) I_t^{\tn{EB-UCB}}\left(A_t^{\tn{BAM}}\right) }\right]
    \leq &\E \left[\sum_{t=1}^T \frac{\widehat{\Delta}^2_{t,\zeta_t(\delta)}\left(A_t^{\tn{EB-UCB}}\right)}{\alpha I_t^{\tn{EB-UCB}}\left(A_t^{\tn{EB-UCB}}\right) + (1-\alpha)I_t^{B}\left(A_t^{\tn{EB-UCB}}\right)}\right] \nonumber \\
    \leq & \frac{1}{1-\alpha}\E \left[\sum_{t=1}^T \frac{\widehat{\Delta}^2_{t,\zeta_t(\delta)}\left(A_t^{\tn{EB-UCB}}\right)}{I_t^{\tn{EB-UCB}}\left(A_t^{\tn{EB-UCB}}\right)} \right].
\end{align}

The next couple of steps are similar to the analysis by \citet{kirschner_thesis}. Let $a_t^{\tn{EB-UCB}}$ be the realization of $A_t^{\tn{EB-UCB}}$. From the Sherman-Morrison formula, we obtain
\begin{align*}
\left(\bo{W}_t + \rho(\aebucb)^{-2}\bo{\phi}(\aebucb)\bo{\phi}(\aebucb)^{\top}\right)^{-1} = \bo{W}_t^{-1} \nonumber - \frac{\rho({\aebucb})^{-2} \bo{W}_t^{-1} \bo{\phi}(\aebucb)\bo{\phi}(\aebucb)^{\top} \bo{W}_t^{-1}}{1 + \rho(\aebucb)^{-2} \bo{\phi}(\aebucb)^{\top} \bo{W}_t^{-1} \bo{\phi}(\aebucb)}
\end{align*}
so
\begin{align*}
    \left \| \bo{\phi}(\aebucb) \right \|^2_{\left(\bo{W}_t + \rho(\aebucb)^{-2}\bo{\phi}(\aebucb)\bo{\phi}(\aebucb)^{\top}\right)^{-1}} = 
    \left \| \bo{\phi}(\aebucb) \right \|^2_{\bo{W}_t^{-1}} - \frac{\rho(\aebucb)^{-2} \left \| \bo{\phi}(\aebucb) \right \|^4_{\bo{W}_t^{-1}}}{1 + \rho(\aebucb)^{-2} \left \| \bo{\phi}(\aebucb) \right \|^2_{\bo{W}_t^{-1}}}.
\end{align*}
Thus
\begin{equation*}
    I_t^{\tn{EB-UCB}}\left(a_t^{\tn{EB-UCB}}\right) = \frac{1}{2}\log\left(1 + \rho(\aebucb)^{-2} \left \| \bo{\phi}(\aebucb) \right \|^2_{\bo{W}_t^{-1}}\right).
\end{equation*}
From \eqref{eq:11-basic}, we have $ \left \| \bo{\phi}(\aebucb) \right \|_{\bo{W}_t^{-1}} \leq U^2\gamma^{-1}$. Thus, using the fact that $\log(1+x) \geq \frac{x}{2q}$ for $q \geq 1$ and $x \in [0, q]$ we get
\begin{equation*}
    I_t^{\tn{EB-UCB}}\left(a_t^{\tn{EB-UCB}}\right) \geq \frac{1}{4}\min\{U^{-2}\gamma , \rho(\aebucb)^{-2}\} \left \| \bo{\phi}(\aebucb) \right \|^2_{\bo{W}_t^{-1}}.
\end{equation*}
So 
\begin{align}\label{eq:Psi_bound_1}
    \frac{\widehat{\Delta}^2_{t,\zeta_t(\delta)}\left(a_t^{\tn{EB-UCB}}\right)}{I_t^{\tn{EB-UCB}}\left(a_t^{\tn{EB-UCB}}\right)} \leq & \frac{4  \beta_{t, \zeta_t(\delta)}(\widehat{B}_t)\left \| \bo{\phi}(\aebucb) \right \|^2_{\bo{W}_t^{-1}}}{\frac{1}{4}\min\{U^{-2}\gamma , \rho(\aebucb)^{-2}\} \left \| \bo{\phi}(\aebucb) \right \|^2_{\bo{W}_t^{-1}}} \nonumber \\
    = & 16 \beta_{t, \zeta_t(\delta)}(\widehat{B}_t) \max \{U^2 \gamma^{-1}, \rho(\aebucb)^2\} \nonumber \\
    \leq & 16 {\beta}_{t, \zeta_t(\delta)}(B) \max \{U^2 \gamma^{-1}, \rho(\aebucb)^2\} \nonumber \\
    \leq &  16 \beta_{T, \zeta_T(\delta)}(B) \max \{U^2 \gamma^{-1}, \rho_{\max}^2\}.
\end{align}
So 
\begin{align}\label{eq:bound_expected_Psi_EB_UCB}
    \E \left[\sum_{t=1}^T \frac{\widehat{\Delta}^2_{t,\zeta_t(\delta)}\left(A_t^{\tn{EB-UCB}}\right)}{I_t^{\tn{EB-UCB}}\left(A_t^{\tn{EB-UCB}}\right)} \right] \leq 16 T \beta_{T, \zeta_T(\delta)}(B) \max \{U^2 \gamma^{-1}, \rho_{\max}^2\}.
\end{align}

Since $1/\zeta_T(\delta) = \max\{1/\delta, T^2\}$, from \eqref{eq:bound_log_det_frac} we have 
\begin{align}\label{eq:bound_beta_T}
    \beta_{T, \zeta_T(\delta)}(B) = & \left(\sqrt{2 \log \left(1 / \zeta_t(\delta)\right)+\log \left(\frac{\operatorname{det}\left(\bo{W}_t\right)}{\operatorname{det}\left(\bo{W}_1\right)}\right)}+\sqrt{\gamma} B\right)^2  \nonumber \\
    \leq  &2 \max\{2\log T, \log(1/\delta) \} + 2\log\left(\frac{\det \bo{W}_T}{\det \bo{W}_1}\right) + 2\gamma B^2 \nonumber \\
 \leq  &2\max \{ 2\log T, \log(1/\delta) \} +  2 d \log(T-1) +2 d \log\left(1 + \frac{\ru}{\gamma}\right) +  2\gamma B^2  \nonumber \\
 <  & (2d + 4)\log T + 2\log(1/\delta) +2 d \log\left(1 + \frac{\ru}{\gamma}\right) +  2\gamma B^2. 
\end{align}
So from \eqref{eq:Psi_exp_1}, \eqref{eq:bound_expected_Psi_EB_UCB}, and \eqref{eq:bound_beta_T} we have
\begin{align}\label{eq:first_eq}
    \E \left[\sum_{t=1}^T \frac{\widehat{\Delta}^2_{t,\zeta_t(\delta)}\left(A_t^{\tn{BAM}}\right)}{\alpha I_t^{B}\left(A_t^{\tn{BAM}}\right) + (1-\alpha) I_t^{\tn{EB-UCB}}\left(A_t^{\tn{BAM}}\right) }\right]
    \leq & \frac{16}{1-\alpha} \max \{U^2 \gamma^{-1}, \rho_{\max}^2\}T  \nonumber \\
    \times  \left[ \vphantom{\frac{1}{2}} (2d + 4)\log T + 2\log(1/\delta) \phantom{a} + \right . & \left. 2 d \log\left(1 + \frac{\ru}{\gamma}\right) +  2\gamma B^2 \right].
\end{align}

Also, for any sequence $\{a_t\}_{t=1}^T \subset \A$ we have 
\begin{align}\label{eq:bound-info-expr}
     I_t^B(a_t) =& \frac{1}{2}\log \left( \| \bo{v}_t^{\min} \|_{(\bo{W}_t + \rho(a)^{-2} \bo{\phi}(a) \bo{\phi}(a)^{\top})}^2\right) - \frac{1}{2}\log \left( \lamin_t) \right) \nonumber \\
     =& \frac{1}{2} \log\left(\frac{(\bo{v}_t^{\min})^{\top}\left(\bo{W}_t + \rho(a_t)^{-2}\bo{\phi}(a_t) \bo{\phi}(a_t)^{\top}\right)\bo{v}_t^{\min}}{\lamin_t)}\right) \nonumber \\
    =& \frac{1}{2} \log\left( \frac{\lamin_t) + \rho(a_t)^{-2}\bo{v}_t^{\min}\bo{\phi}(a_t)\bo{\phi}(a_t)^{\top} \bo{v}_t^{\min}}{\lambda_{\min}(\bo{W}_t)}\right)\nonumber  \\
    =& \frac{1}{2}\log\left(1 + \frac{\rho(a_t)^{-2} {\psi}_{\bo{v}_t^{\min}}(\bo{\phi}(a_t))^2}{\lambda_{\min}(\bo{W}_t)}\right) = \frac{1}{2}\log\left(1 + \frac{\omega_t(a_t)}{\lambda_{\min}(\bo{W}_t)}\right).
\end{align}

Let 
$$T_0 = \max \left\{t : \sum_{\tau =1}^{t}\omega_{\tau}(a_{\tau}) \leq d(\ru - \gamma)\right\}.$$
Without loss of generality, assume that $T_0 \leq T$. Then using Lemma \ref{lem:eigvals} we get 
\begin{align*}
    \sum_{t=1}^T I_t^B(a_t) = &\sum_{t=1}^T \log\left(1 + \frac{\omega_t(a_t)}{\lambda_{\min}(\bo{W}_t)}\right) \\
    =& \sum_{t=1}^{T_0} \log \left(1 + \frac{\omega_t(a_t)}{\lambda_{\min}(\bo{W}_t)}\right) + \sum_{t=T_0 + 1}^{T} \log \left(1 + \frac{\omega_t(a_t)}{\lambda_{\min}(\bo{W}_t)}\right)
    \\
     \leq & \sum_{t=1}^{T_0} \frac{\omega_t(a_t)}{\lambda_{\min}(\bo{W}_t)} + \sum_{t=T_0 + 1}^{T} \log \left(1 + \frac{\omega_t(a_t)}{\gamma - \ru + \frac{1}{d} \sum_{\tau=1}^{t-1} \omega_{\tau}(a_{\tau})}\right) 
    \\
    \leq & \frac{1}{\gamma}\sum_{t=1}^{T_0} \omega_{t}(a_{t}) + \sum_{t=T_0 + 1}^{T} \log \left(1 + \frac{d\omega_t(a_t)}{d(\gamma - \ru) + \sum_{\tau=1}^{t-1} \omega_{\tau}(a_{\tau})}\right) 
    \\
    \leq & \frac{d(\ru - \gamma)}{\gamma} + \sum_{t=T_0 + 1}^{T} \frac{d\omega_t(a_t)}{d(\gamma - \ru) + \sum_{\tau=1}^{T_0} \omega_{\tau}(a_{\tau}) + \sum_{\tau=T_0 + 1}^{t-1} \omega_{\tau}(a_{\tau})}.
\end{align*}
Let 
$$
c = d(\gamma - \ru) + \sum_{\tau=1}^{T_0} \omega_{\tau}(a_{\tau})
$$
and
$$
x_t = \omega_{T_0 + t}(a_{T_0 + t}).
$$
Then from Lemma \ref{lem:sequences}, since $c > 0$ and $x_t \in [0, \ru]$ for all $t$ we have
\begin{align}\label{eq:bound_B_info}
    \sum_{t=1}^T I_t^B(a_t) \leq & \frac{d(\ru - \gamma)}{\gamma} + \sum_{t=T_0 + 1}^{T} \frac{d\omega_t(a_t)}{c + \sum_{\tau=T_0 + 1}^{t-1} \omega_{\tau}(a_{\tau})} \nonumber \\ 
    = & \frac{d(\ru - \gamma)}{\gamma} + d \sum_{t=1}^{T-T_0} \frac{x_t}{c + \sum_{\tau=1}^{t-1} x_{\tau}} \nonumber \\
     \leq & O(d \log(T - T_0)) \leq O(d \log T).
\end{align}
Thus, from \eqref{eq:bound_UCB_info} we have
\begin{align*}
    \alpha \E \left[ \sum_{t=1}^T I_t^{B}\left(A_t^{\tn{BAM}}\right)\right] + (1-\alpha) \E \left[\sum_{t=1}^T I_t^{\tn{EB-UCB}}\left(A_t^{\tn{BAM}}\right)\right] \leq O(d \log T).
\end{align*}

So from Theorem \ref{thm:univ_IDS_bound} and \eqref{eq:first_eq} we have 
\begin{align}\label{eq:Delta_hat_1_bound}
    \E \left[ \sum_{t=1}^T \widehat{\Delta}_{t,\zeta_t(\delta)} (A_t^{\textnormal{BAM}})\right] \leq & 
   O\left( \sqrt{d\frac{16}{1-\alpha} \max \{U^2 \gamma^{-1}, \rho_{\max}^2\} T \log T} \right. \nonumber \\
   &\times \left. \sqrt{(4 + 2d)\log T + 2\log(1/\delta) +2 d \log\left(1 + \frac{\ru}{\gamma}\right) +  2\gamma B^2 }\right) \nonumber\\
  \leq &  O\left(\frac{d \max \{U / \sqrt{\gamma}, \rho_{\max}\}}{\sqrt{1-\alpha}} \sqrt{T} \log T \right. \nonumber \\ 
  &\times \left. \sqrt{ \log (1/\delta) + \log\left(1 + \frac{\ru}{\gamma}\right) + \gamma B^2 }  \right).
\end{align}

Take any $t \geq 1$ and suppose that the event $E_{t, \zeta_t(\delta)}$, as defined in \eqref{eq:event_E_t}, holds. Note that the set 
$$
\left\{ \bo{\theta} \in \R : \left\| \bo{\theta} - \widehat{\bo{\theta}}_t^{\mathrm{wls}} \right \|_{\bo{W}_t}^2 \leq \beta_{t,\zeta_t(\delta)}(B^*)\right\}
$$         
is an ellipsoid in $\R^d$ centered at $\widehat{\bo{\theta}}_t^{\mathrm{wls}}$ with the longest semi-axis of length $\beta_{t, \zeta_t(\delta)}^{{1}/{2}}(B^*)\lambda_{\min}(\bo{W}_t)^{-{1}/{2}}$, so 
\begin{equation}\label{eq:1}
    \left\|\widehat{\bo{\theta}}_t^{\mathrm{wls}} - \bo{\theta}^* \right\|_2 \leq \beta_{t, \zeta_t(\delta)}^{ {1}/{2}}(B^*)\lambda_{\min}(\bo{W}_t)^{-{1}/{2}}.
\end{equation}
Since $B \geq B^*$ we have ${\beta}_{t, \zeta_t(\delta)}(B) \geq \beta_{t, \zeta_t(\delta)}(B^*)$,
so by the triangle inequality we get
\begin{equation}\label{eq:hat_B_star_B}
    B^* = \| \bo{\theta}^* \|_2 \leq  \left\| \widehat{\bo{\theta}}_t^{\mathrm{wls}} \right\|_2 +\beta^{ 1/2}_{t, \zeta_t(\delta)}(B^*) \lambda_{\min}(\bo{W}_t)^{-{1}/{2}} \leq \left\| \widehat{\bo{\theta}}_t^{\mathrm{wls}} \right\|_2 + \beta_{t, \zeta_t(\delta)}^{1/2}(B) \lambda_{\min}(\bo{W}_t)^{-{1}/{2}} = \widehat{B}_t.
\end{equation}
So $B^* \leq \widehat{B}_t$
for all $t \geq 1$ and thus $\beta_{t, \zeta_t(\delta)}(\widehat{B}_t) \geq \beta_{t, \zeta_t(\delta)}(B^*)$, so
\begin{equation*}
    \bo{\theta}^* \in \left\{ \bo{\theta} \in \R : \left\| \bo{\theta} - \widehat{\bo{\theta}}_t^{\mathrm{wls}} \right \|_{\bo{W}_t}^2 \leq \beta_{t, \zeta_t(\delta)}(B^*)\right\} \subseteq \left\{ \bo{\theta} \in \R : \left\| \bo{\theta} - \widehat{\bo{\theta}}_t^{\mathrm{wls}} \right \|_{\bo{W}_t}^2 \leq \beta_{t, \zeta_t(\delta)}(\widehat{B}_t)\right\}
\end{equation*}
Hence $\Delta(a) \leq \widehat{\Delta}_{t, \zeta_t(\delta)}(a)$ for all $a \in \A$. So for any $a \in \A$ we have
\begin{equation*}
     \Prob \left(\Delta(a) > \widehat{\Delta}_{t, \zeta_t(\delta)}(a)\right) \leq 1 -\Prob(E_{t, \zeta_t(\delta)}) \leq \zeta_t(\delta) \leq 1/t^2.
\end{equation*}

Thus, letting  $\Delta_{\max} = \max_{a \in \A} \Delta (a)$, for any sequence $\{a_t \}_{t=1}^{T} \subset \A$ we have 
\begin{equation}\label{eq:delta_hat_gap}
    \E\left[ \sum_{t=1}^T \Delta{(a_t)} - \widehat{\Delta}_{t, \zeta_t(\delta)}(a_t) \right] \leq \Delta_{\max} \sum_{t=1}^T \Prob\left( \Delta{(a_t)} > \widehat{\Delta}_{t, 1/t^2}(a_t)\right) \leq \Delta_{\max} \sum_{t=1}^T \frac{1}{t^2} \leq  O(\Delta_{\max}).
\end{equation}
So from \eqref{eq:Delta_hat_1_bound}, for $T \leq T_B$ the regret of EBIDS is bounded above by
\begin{equation*}
    \mathcal{R}_T  \leq O\left(\frac{d \max \{U / \sqrt{\gamma}, \rho_{\max}\}}{\sqrt{1-\alpha}} \sqrt{T} \log T \sqrt{ \log (1/\delta) + \log\left(1 + \frac{\ru}{\gamma}\right) + \gamma B^2 }  \right).
\end{equation*}

From \eqref{eq:bound_UCB_info} and \eqref{eq:bound_B_info} with probability $1$ we have
\begin{equation}\label{eq:bound_BAM_info}
    \sum_{t=1}^T I_t^{\tn{BAM}}(\Abam) \leq O(d \log T).
\end{equation}
Following the same steps as in \eqref{eq:Psi_exp_1}, using \eqref{eq:Psi_bound_1} and \eqref{eq:first_eq} we have
\begin{align}\label{eq:Psi_BAM_bound}
  \sum_{t=1}^T \Psi_t^{\textnormal{BAM}}\left(A_t^{\tn{BAM}}\right) = &  \sum_{t=1}^T \frac{\widehat{\Delta}^2_{t,\zeta_t(\delta)}\left(A_t^{\tn{BAM}}\right)}{I_t^{\tn{BAM}}\left(A_t^{\tn{BAM}}\right)} \leq \sum_{t=1}^T \frac{\widehat{\Delta}^2_{t,\zeta_t(\delta)}\left(A_t^{\tn{BAM}}\right)}{\alpha I_t^{B}\left(A_t^{\tn{BAM}}\right) + (1-\alpha) I_t^{\tn{EB-UCB}}\left(A_t^{\tn{BAM}}\right) }  \nonumber \\
   \leq & \sum_{t=1}^T   \frac{\widehat{\Delta}^2_{t,\zeta_t(\delta)}\left(A_t^{\tn{EB-UCB}}\right)}{\alpha I_t^{\tn{EB-UCB}}\left(A_t^{\tn{EB-UCB}}\right) + (1-\alpha)I_t^{B}\left(A_t^{\tn{EB-UCB}}\right)} \nonumber \\
   \leq & 
    \frac{1}{1-\alpha}\sum_{t=1}^T \frac{\widehat{\Delta}^2_{t,\zeta_t(\delta)}\left(A_t^{\tn{EB-UCB}}\right)}{I_t^{\tn{EB-UCB}}\left(A_t^{\tn{EB-UCB}}\right)} \nonumber \\
   \leq & \frac{16}{1-\alpha}  \sum_{t=1}^T {\beta}_{T, \zeta_T(\delta)}(B) \max \{U^2 \gamma^{-1}, \rho_{\max}^{2}\} \nonumber \\
   \leq & \frac{16}{1-\alpha} \max \{U^2 \gamma^{-1}, \rho_{\max}^2\} T \nonumber \\
   &\times \left[ (4+2d)\log T + 2\log(1/\delta) + 2 d \log\left(1 + \frac{\ru}{\gamma}\right) +  2\gamma B^2 \right].
\end{align}

Following analogous steps as above, since $1/\zeta_t(\delta) \geq 1/ \delta$ we have $\beta_{t, \zeta_t(\delta)}(B) \geq \beta_{t, \delta}(B) \geq \beta_{t, \delta}(B^*)$. So for any $t\geq 1$, whenever event $E_{t, \delta}$ holds, the inequality $B^* \leq \widehat{B}_t$ holds as well and thus $\Delta(a) \leq \widehat{\Delta}_{t, \zeta_t(\delta)}(a)$, for all $a \in \A$. So if $E_{\delta} = \bigcap_{t=1}^{\infty}E_{t,\delta}$ holds, then $\Delta(a) \leq \widehat{\Delta}_{t, \zeta_t(\delta)}(a)$, for all $a \in \A$ and for all $t \geq 1$. 
So from \eqref{eq:Psi_BAM_bound} and \eqref{eq:bound_BAM_info}, by Theorem \ref{thm:univ_IDS_bound} we have
\begin{equation*}
    \mathcal{PR}_T  \leq O\left(\frac{d \max \{U / \sqrt{\gamma}, \rho_{\max}\}}{\sqrt{1-\alpha}} \sqrt{T} \log T \sqrt{ \log (1/\delta) + \log\left(1 + \frac{\ru}{\gamma}\right) + \gamma B^2 }  \right). \qed
\end{equation*}

\subsection{Proof of Proposition \ref{prop:high_prob_B}} \label{proof_high_prob_B}

In order to precisely state the conditions on $T_B$ and $\alpha$, i.e., how large each of them needs to be for $B^* \leq \tilde{B}_t \leq (1+g)B^*$ to hold for all $t \geq T_B+1$, we will first define several constants for notational convenience.

Let 
\begin{equation}\label{eq:c_0}
c_0 = L^2\left[U^2(\gamma + \ru)\left(\frac{1}{\kappa} + \frac{1}{\gamma}\right)\right]^{-1}
\end{equation}
and
\begin{equation}\label{eq:h_0}
     h_0 = 8 \log (5/4) + 4\log(1/\delta) +2d\log\left(1+\frac{\ru}{\gamma}\right) + 2\gamma B^2.
\end{equation}
Then let 
\begin{align}
    u_0 &= \frac{c_0}{6+16g^{-2}}\log 2 + \frac{1-\alpha}{\alpha} d \log\left (1 + \frac{\ru}{\gamma}\right) \label{eq:u_0} \\
    u_1 &= \frac{c_0}{12+32g^{-2}} - \frac{1-\alpha}{2\alpha}d  \label{eq:u_1} \\
    w_0 &= \frac{c_0}{6+16g^{-2}} + \frac{1-\alpha}{\alpha} d \log \left(1 + \frac{\ru}{\gamma}\right) \label{eq:w_0} \\
   w_1  &= \frac{c_0}{12+32g^{-2}} - \frac{1-\alpha}{\alpha} d. \label{eq:w_1}
\end{align}

and finally let
\begin{align}
    b_0 &= \frac{1}{d}\left[w_0\left(\frac{\gamma}{d} u_0 - \gamma + \ru\right) - \gamma u_0 \right] + \gamma - \ru \label{eq:b_0} \\
    b_1 &= \frac{1}{d}\left(\gamma u_1 - \frac{\gamma}{d}u_1w_0 - \frac{\gamma}{d} u_0 w_1 + \gamma w_1 - \ru w_1 \right) \label{eq:b_1} \\
    b_2 &=  \frac{\gamma}{d^2} u_1 w_1 \label{eq:b_2}
\end{align}
We make the following assumptions.

\begin{assumption}\label{as:B}
$B \geq B^*$.
\end{assumption}

\begin{assumption} \label{as:T_B}
$$
T_B \geq \max\left\{4, \exp\left[\frac{h_0 + 2d + 8}{b_2}\left(4g^{-2}B^{* -2} + \frac{|b_1|}{2d +8} + \frac{|b_0|}{h_0 + 2d + 8}\right)\right]\right\}.
$$
\end{assumption}

\begin{assumption}\label{as:alpha}
\[
\alpha \geq \frac{d}{d + \frac{c_0}{12+32g^{-2}}}.
\]

\end{assumption}

We will now show that if Assumptions \ref{as:B}-\ref{as:alpha} are satisfied and event $E_{\delta}$ holds then $B^* \leq \tilde{B}_t \leq (1+g)B^*$ for all $t \geq T_B+1$.

\proof
Suppose that event $E_{\delta}$ holds. 
 For any $t$ let
 \begin{equation}\label{eq:s_def}
     {s}(t) = \argmin_{\tau\leq t} {\beta}^{1/2}_{\tau, \zeta_{\tau}(\delta)}(\widehat{B}_{\tau}) \lambda_{\min}(\bo{W}_{\tau})^{-1/2}
 \end{equation}
From \eqref{eq:1} in the proof of Proposition \ref{prop:regret_1}, using the triangle inequality we get 
\begin{align}\label{eq:2}
   \left\| \widehat{\bo{\theta}}_{t}^{\mathrm{wls}} \right\|_2 \leq & \left\| {\bo{\theta}}^* \right\|_2  + \beta_{t,\zeta_{t}(\delta)}^{ {1}/{2}}(B^*)\lambda_{\min}(\bo{W}_{t})^{-{1}/{2}} =  B^*   + \beta_{t, \zeta_{t}(\delta)}^{{1}/{2}}(B^*)\lambda_{\min}(\bo{W}_{t})^{-{1}/{2}}.
\end{align}
From \eqref{eq:hat_B_star_B} in the proof of Proposition \ref{prop:regret_1}, for any $t$ we have $\widehat{B}_{t} \geq B^*$, so 
\begin{align*}
    \left\| \widehat{\bo{\theta}}_{t}^{\mathrm{wls}} \right\|_2 \leq B^* + \beta_{t, \zeta_{t}(\delta)}^{{1}/{2}}(\widehat{B}_{t})\lambda_{\min}(\bo{W}_{t})^{-{1}/{2}}.
\end{align*}

Hence
\begin{align}\label{eq:4_star_star}
    \tilde{B}_t = & \min_{\tau \leq t}\left \{ \left\| \widehat{\bo{\theta}}_{\tau}^{\mathrm{wls}} \right\|_2 + {\beta}^{1/2}_{\tau,\zeta_{\tau}(\delta)}(\widehat{B}_{\tau}) \lamin_{\tau})^{-1/2} \right \} \nonumber \\
    \leq & \left\| \widehat{\bo{\theta}}_{s(t)}^{\mathrm{wls}} \right\|_2 + {\beta}^{1/2}_{s(t),\zeta_{s(t)}(\delta)}(\widehat{B}_{s(t)}) \lamin_{s(t)})^{-1/2}  \nonumber \\
   \leq & B^* + 2 {\beta}^{1/2}_{s(t),\zeta_{s(t)}(\delta)}(\widehat{B}_{s(t)})\lamin_{s(t)})^{-1/2}.
\end{align}
Also, analogously as in \eqref{eq:hat_B_star_B}, using \eqref{eq:1} and the triangle inequality, for any $t \geq 1$ we have
\begin{align*}
    B^* = \left\| \bo{\theta}^* \right\|_2 \leq \left\| \widehat{\bo{\theta}}_{t}^{\mathrm{wls}} \right\|_2 + {\beta}^{ 1/2}_{t,\zeta_{t}(\delta)}(B^*) \lamin_{t})^{-1/2} \leq 
    \left\| \widehat{\bo{\theta}}_{t}^{\mathrm{wls}} \right\|_2 + {\beta}^{1/2}_{t,\zeta_{t}(\delta)}(\widehat{B}_{t}) \lamin_{t})^{-1/2}.
\end{align*}
So 
\begin{equation}\label{eq:4_star_star_star}
B^* \leq \tilde{B}_t
\end{equation}
for any $t \geq 1$.

From Lemma \ref{lem:ids_mixtures}, for any $t \leq T_B$ we have
\begin{equation}\label{4_plus}
    I_t^B(a_t^{\textnormal{BAM}}) \geq \frac{\widehat{\Delta}^2_{t, \zeta_t(\delta)}(a_t^{\textnormal{BAM}})}{\widehat{\Delta}^2_{t, \zeta_t(\delta)}\left(a_t^{{I}, B}\right)} I_t^B\left(a_t^{I, B}\right) - \frac{1-\alpha}{\alpha}I_t^{\textnormal{EB-UCB}}\left(a_t^{\textnormal{BAM}}\right)
\end{equation}
where $a_t^{I,B} = \argmax_{a \in \A} I_t^B(a)$.
For any $t \leq T_B$ we have
\begin{align*}
    \widehat{\Delta}_{t, \zeta_t(\delta)}\left(a_t^{\textnormal{BAM}}\right)
    =& \max_{b \in \A} \left\{ \left\langle \bo{\phi}(b), \widehat{\bo{\theta}}^{\mathrm{wls}}_t \right\rangle  + {\beta}_{t, \zeta_t(\delta) }^{1/2}(\widehat{B}_t)\| \bo{\phi}(b) \|_{\bo{W}_t^{-1}}\right\} \\
    &-\left( \left\langle \bo{\phi}\left(\abam\right), \widehat{\bo{\theta}}^{\mathrm{wls}}_t \right\rangle - \beta_{t, \zeta_t(\delta)}^{1/2}(\widehat{B}_t)\left \|\bo{\phi}\left(\abam\right) \right\|_{\bo{W}_t^{-1}} \right) \\
    =& \max_{b \in \A} \left\{ \left\langle \bo{\phi}(b), \widehat{\bo{\theta}}^{\mathrm{wls}}_t \right\rangle  + {\beta}_{t, \zeta_t(\delta) }^{1/2}(\widehat{B}_t)\| \bo{\phi}(b) \|_{\bo{W}_t^{-1}}\right\} \\
     &- \left( \left\langle \bo{\phi}\left(\abam\right), \widehat{\bo{\theta}}^{\mathrm{wls}}_t \right\rangle + \beta_{t, \zeta_t(\delta)}(\widehat{B}_t)^{1/2}\left \|\bo{\phi}\left(\abam\right) \right\|_{\bo{W}_t^{-1}} \right) \\
     & + 2 \beta_{t, \zeta_t(\delta)}^{1/2}(\widehat{B}_t)\left \|\bo{\phi}\left(\abam\right) \right\|_{\bo{W}_t^{-1}} \\
    \geq & 2 \beta_{t, \zeta_t(\delta)}^{1/2}(\widehat{B}_t)\left \|\bo{\phi}\left(\abam\right) \right\|_{\bo{W}_t^{-1}}.
\end{align*}
So from \eqref{eq:13-basic} 
\begin{align}\label{eq:roman_1}
    \widehat{\Delta}^2_{t, \zeta_t(\delta)}\left(a_t^{\textnormal{BAM}}\right) \geq 4 \beta_{t, \zeta_t(\delta)}(\widehat{B}_t)\left \|\bo{\phi}\left(\abam\right) \right\|_{\bo{W}_t^{-1}}^2 
     \geq 4\beta_{t, \zeta_t(\delta)}(\widehat{B}_t)\frac{L^2}{t(\gamma + \ru)}
\end{align}
Also 
\begin{align*}
    \widehat{\Delta}^2_{t, \zeta_t(\delta)}\left(a_t^{I,B}\right) =& \beta^{1/2}_{t, \zeta_t(\delta)}(\widehat{B}_t) \left(\left\| \bo{\phi}(a_t^{\textnormal{EB-UCB}})\right\|_{\bo{W}_t^{-1}} + \left\| \bo{\phi}(a_t^{I,B})\right\|_{\bo{W}_t^{-1}}\right) \nonumber \\
    & + \left\langle \bo{\phi}(a_t^{\textnormal{EB-UCB}}), \widehat{\bo{\theta}}^{\mathrm{wls}}_t\right\rangle - \left\langle \bo{\phi}(a_t^{I,B}), \widehat{\bo{\theta}}^{\mathrm{wls}}_t\right\rangle,
\end{align*}
so
\begin{align*}
    \widehat{\Delta}^2_{t, \zeta_t(\delta)}\left(a_t^{I,B}\right)
    \leq & 4\beta_{t, \zeta_t(\delta)}(\widehat{B}_t) \left(\left\| \bo{\phi}(a_t^{\textnormal{EB-UCB}})\right\|^2_{\bo{W}_t^{-1}} + \left\| \bo{\phi}(a_t^{I,B})\right\|^2_{\bo{W}_t^{-1}}\right) \nonumber \\
    &+ 4\left\langle \bo{\phi}(a_t^{\textnormal{EB-UCB}}), \widehat{\bo{\theta}}^{\mathrm{wls}}_t\right\rangle^2 +  4\left\langle \bo{\phi}(a_t^{I,B}), \widehat{\bo{\theta}}^{\mathrm{wls}}_t\right\rangle^2.
\end{align*}
Since $E_{\delta}$ holds, from \eqref{eq:star_cauchy_schwarz} and \eqref{eq:2} for any $t$ and any $a \in \A$ we have 
\begin{align*}
    \left\langle \bo{\phi}(a), \widehat{\bo{\theta}}_t^{\mathrm{wls}} \right\rangle^2 \leq 2U^2(B^{* 2} + \beta_{t, \zeta_t(\delta)} (B^*) \lamin_t)^{-1})
\end{align*}
so from \eqref{eq:11-basic} we have 
\begin{align}\label{eq:7}
    \widehat{\Delta}_{t, \zeta_t(\delta)}^2\left(a_t^{I,B}\right) \leq 8\beta_{t, \zeta_t(\delta)}(\widehat{B}_t) U^2\lamin_t)^{-1} + 16U^2(B^{* 2} + \beta_{t, \zeta_t(\delta)}(B^*)\lamin_t)^{-1}).
\end{align}
From \eqref{eq:bound-info-expr} from the proof of Proposition \ref{prop:regret_1}, for any $a \in \A$ we have
\begin{align}\label{eq:dagger-3}
    I_t^B(a) = \frac{1}{2}\log\left(1 + \frac{\rho(a)^{-2}\psi_{\bo{v}_t^{\min}}\left(\bo{\phi}(a)\right)^2}{\lamin_t)}\right),
\end{align}
so
\begin{align*}
    I_t^B\left(a_t^{I,B}\right) = & \max_{a \in \A}I_t^B(a) = \max_{a \in \A}\left\{\frac{1}{2}\log\left(1 + \frac{\rho(a)^{-2}\psi_{\bo{v}_t^{\min}}\left(\bo{\phi}(a)\right)^2}{\lamin_t)}\right) \right\} \nonumber \\
    \geq & \frac{1}{2} \log\left(1 + \frac{\kappa}{\lamin_t)}\right).
\end{align*}
Thus, since $\log x \geq 1 - \frac{1}{x}$ for all $x > 0$, we have
\begin{align}\label{eq:8}
    I_t^{B}\left(a_t^{I,B}\right) \geq & \frac{\kappa}{2(\lamin_t) + \kappa)} = \left[2\lamin_t)\left(\frac{1}{\kappa} + \frac{1}{\lamin_t)}\right)\right]^{-1} \nonumber \\
    \geq & \left[2\lamin_t)\left(\frac{1}{\kappa} + \frac{1}{\gamma}\right)\right]^{-1}.
\end{align}

So combining \eqref{4_plus}, \eqref{eq:roman_1}, \eqref{eq:7}, and \eqref{eq:8}, for any $t\leq T_B$ we have
\begin{align*}
    I_t^B(\abam) \geq & \frac{L^2 \lamin_t)^{-1} \left[2t (\gamma+\ru)\left(\frac{1}{\kappa} + \frac{1}{\gamma}\right)\right]^{-1}}{2U^2\lamin_t)^{-1} + 4U^2 \beta_{t, \zeta_t(\delta)}(\widehat{B}_t)^{-1}\left(B^{* 2} + \beta_{t, \delta}(B^*)\lamin_t)^{-1}\right)} \nonumber \\
    &- \frac{1-\alpha}{\alpha}I_t^{\textnormal{EB-UCB}}\left( \abam\right) = \\
     = & \frac{1}{t}L^2\left[U^2(\gamma + \ru)\left(\frac{1}{\kappa} + \frac{1}{\gamma}\right)\left(4 + 8B^{* 2}\frac{\lamin_t)}{\beta_{t, \zeta_t(\delta)}(\widehat{B}_t)} + 8\frac{\beta_{t, \delta}(B^*)}{\beta_{t, \zeta_t(\delta)}(\widehat{B}_t)}\right)\right]^{-1} \nonumber \\
     & - \frac{1-\alpha}{\alpha}I_t^{\textnormal{EB-UCB}}(\abam) \geq \\
     \geq & \frac{1}{t}L^2\left[U^2(\gamma + \ru)\left(\frac{1}{\kappa} + \frac{1}{\gamma}\right)\left(12 + 8B^{* 2}\frac{\lamin_t)}{\beta_{t, \zeta_t(\delta)}(\widehat{B}_t)}\right)\right]^{-1} \nonumber \\
    &- \frac{1-\alpha}{\alpha}I_t^{\textnormal{EB-UCB}}(\abam),
\end{align*}
where the last inequality follows from the fact that $\widehat{B}_t \geq B^*$ and $1/\zeta_t(\delta) \geq 1/\delta$ which gives us
\begin{align*}
    \frac{\beta_{t, \delta}(B^*)}{\beta_{t, \zeta_t(\delta)}(\widehat{B}_t)} \leq 1.
\end{align*}

So from \eqref{eq:c_0} we have 
\begin{align}\label{eq:9_star}
    I_t^B(\abam) \geq \frac{1}{t}c_0\left(12 + 8 B^{* 2}\frac{\lamin_t)}{\beta_{t, \zeta_t(\delta)}(\widehat{B}_t)}\right)^{-1} - \frac{1-\alpha}{\alpha}I_t^{\textnormal{EB-UCB}}(\abam).
\end{align}
From \eqref{eq:dagger-3} we have
\begin{align*}
     I_t^B(\abam) = \frac{1}{2}\log\left(1 + \frac{\omega_t(\abam)}{\lamin_t)}\right) \leq  \frac{\omega_t(\abam)}{2\lamin_t)}.
\end{align*}
So 
\begin{align}\label{eq:9_star_star}
    \omega_t(\abam) \geq 2\lamin_t) I_t^B(\abam).
\end{align}

If 
\begin{equation}\label{eq:10}
    \beta_{t, \zeta_t(\delta)}^{1/2}(\widehat{B}_t) \lamin_t)^{-1/2} \leq \frac{1}{2}gB^*
\end{equation}
 for some $t\leq T_B+1$ then
 \begin{equation*}
     {\beta}^{1/2}_{s(t), \zeta_{s(t)}(\delta)}(\widehat{B}_{s(t)})\lamin_{s(t)})^{-1/2} \leq \frac{1}{2}g B^*,
 \end{equation*}
 so from \eqref{eq:4_star_star} and \eqref{eq:4_star_star_star}, since event $E_{\delta}$ holds, for any $t \geq T_B+1$ we have
\begin{equation}\label{eq:11}
    B^* \leq \tilde{B}_t \leq B^* + 2{\beta}^{1/2}_{s(t), \zeta_{s(t)}(\delta)}(\widehat{B}_t)\lamin_{s(t)})^{-1/2} = (1+g)B^*
\end{equation}
which is what we want to show.
We will prove by contradiction that since $E_{\delta}$ holds, \eqref{eq:10} holds as well for some $t\leq T_B+1$. Suppose that \eqref{eq:10} does not hold. Then for all $t \leq T_B + 1$ we have
\begin{equation}\label{eq:12}
    \frac{\lamin_t)}{\beta_{t, \zeta_t(\delta)}(\widehat{B}_t)} < 4g^{-2}B^{* -2},
\end{equation}
so from \eqref{eq:9_star} we have 
\begin{align*}
    I_t^B(\abam) \geq & \frac{1}{t}c_0\left(12 + 8 B^{* 2}\frac{\lamin_t)}{\beta_{t, \zeta_t(\delta)}(\widehat{B}_t)}\right)^{-1} - \frac{1-\alpha}{\alpha}I_t^{\textnormal{EB-UCB}}(\abam) \nonumber \\
    > & \frac{1}{t}\cdot \frac{c_0}{12+32g^{-2}} -  \frac{1-\alpha}{\alpha}I_t^{\textnormal{EB-UCB}}(\abam).
\end{align*}
Hence, from \eqref{eq:9_star_star} for any $t\leq T_B$ we have
\begin{equation*}
    \omega_t(\abam) \geq \lamin_t)\left(\frac{1}{t}\cdot \frac{c_0}{6+16g^{-2}} - 2 \frac{1-\alpha}{\alpha}I_t^{\textnormal{EB-UCB}}(\abam)\right).
\end{equation*}

Let $\lfloor x \rfloor$ denote the largest integer smaller than or equal to $x$ for any $x \in \R$. From Weyl's inequality \citep{matrix_theory}
\begin{equation}
\label{eq:lambda_seq_increasing}
\lamin_{t+1}) \geq \lamin_t) \geq \gamma
\end{equation}
for any $t$. Also note that $\omega_t(a) \geq 0$ for any $t$ and any $a \in \A$. So
\begin{align*}
    \sum_{t=1}^{\lfloor \sqrt{T_B}\rfloor} \omega_t(\abam) \geq \gamma \left(\frac{c_0}{6 + 16g^{-2}}\sum_{t=1}^{\lfloor \sqrt{T_B}\rfloor}\frac{1}{t}- 2\frac{1-\alpha}{\alpha} \sum_{t=1}^{\lfloor \sqrt{T_B}\rfloor} I_t^{\textnormal{EB-UCB}}(\abam)\right).
\end{align*}

From \eqref{eq:bound_UCB_info} we have
\begin{align*}
    \sum_{t=1}^{\lfloor \sqrt{T_B}\rfloor} I_t^{\textnormal{EB-UCB}}(\abam) \leq & \frac{1}{2} d\log\lfloor \sqrt{T_B} \rfloor + \frac{1}{2} d \log \left(1 + \frac{\ru}{\gamma}\right) \nonumber \\
    \leq & \frac{1}{4} d\log T_B + \frac{1}{2} d \log \left(1 + \frac{\ru}{\gamma}\right).
\end{align*}
Also since $T_B \geq 4$ we have $\lfloor \sqrt{T_B} \rfloor \geq \sqrt{T_B} - 1 \geq \sqrt{T_B} /2$, so
\begin{equation*}
    \sum_{t=1}^{\lfloor \sqrt{T_B}\rfloor} \frac{1}{t} > \log \lfloor \sqrt{T_B} \rfloor \geq \log\left(\frac{1}{2}\sqrt{T_B}\right) = \frac{1}{2}\log T_B - \log 2.
\end{equation*}
So 
\begin{align}\label{eq:14}
\sum_{t=1}^{\lfloor \sqrt{T_B}\rfloor} \omega_t(\abam) \geq & \gamma \left(\frac{c_0}{6+16g^{-2}}\left[\frac{1}{2}\log T_B - \log 2\right] - \frac{1-\alpha}{\alpha}d \left[\frac{1}{2}\log T_B + \log \left( 1 + \frac{\ru}{\gamma}\right)\right]\right) \nonumber \\
\geq &  \gamma\left(\left[\frac{c_0}{12+32g^{-2}} - \frac{1-\alpha}{2\alpha}d\right]\log T_B - \left[\frac{c_0}{6+16g^{-2}}\log 2 + \frac{1-\alpha}{\alpha} d \log\left (1 + \frac{\ru}{\gamma}\right)\right]\right) \nonumber \\
= & \gamma(u_1 \log T_B - u_0),
\end{align}
where the constants $u_0$ and $u_1$ were defined in \eqref{eq:u_0} and \eqref{eq:u_1}, respectively.
Similarly from \eqref{eq:lambda_seq_increasing} we have
\begin{align*}
    \sum_{t=\lfloor \sqrt{T_B} \rfloor + 1}^{T_B} \omega_t(\abam) \geq \lamin_{\lfloor \sqrt{T_B} \rfloor + 1}) \left(\frac{c_0}{6 + 16g^{-2}}\sum_{t=\lfloor \sqrt{T_B}\rfloor+ 1}^{T_B}\frac{1}{t} - 2\frac{1-\alpha}{\alpha} \sum_{t=\lfloor \sqrt{T_B}\rfloor + 1}^{T_B} I_t^{\textnormal{EB-UCB}}(\abam)\right)
\end{align*}
Note hat 

\begin{equation*}
     \sum_{t=\lfloor \sqrt{T_B} \rfloor + 1}^{T_B}  I_t^{\textnormal{EB-UCB}}(\abam) \leq  \sum_{t= 1}^{T_B}  I_t^{\textnormal{EB-UCB}}(\abam) \leq \frac{1}{2} d\log T_B + \frac{1}{2} d\log \left(1 + \frac{\ru}{\gamma}\right)
\end{equation*}
and
\begin{align*}
    \sum_{t=\lfloor \sqrt{T_B} \rfloor + 1}^{T_B} \frac{1}{t} = \sum_{t=1}^{T_B} \frac{1}{t} - \sum_{t=1}^{\lfloor \sqrt{T_B} \rfloor} \frac{1}{t} > \log T_B - (\log \sqrt{T_B} + 1) = \frac{1}{2}\log T_B - 1.
\end{align*}
So
\begin{align*}
    \sum_{t=\lfloor \sqrt{T_B} \rfloor + 1}^{T_B} \omega_t(\abam) \geq & \lamin_{\lfloor \sqrt{T_B} \rfloor + 1}) \nonumber \\
    & \times \left(\frac{c_0}{6+16g^{-2}} \left[\frac{1}{2}\log T_B - 1\right] - \frac{1-\alpha}{\alpha}d\left[\log T_B + \log \left(1 + \frac{\ru}{\gamma}\right)\right]\right) \\
     = & \lamin_{\lfloor \sqrt{T_B} \rfloor + 1}) \nonumber \\ 
     &\times \left( \left[\frac{c_0}{12+32g^{-2}} - \frac{1-\alpha}{\alpha} d\right] \log T_B - \left[\frac{c_0}{6+16g^{-2}} + \frac{1-\alpha}{\alpha} d \log \left(1 + \frac{\ru}{\gamma}\right)\right] \right) \\
     = &  \lamin_{\lfloor \sqrt{T_B} \rfloor + 1})(w_1 \log T_B + w_0),
\end{align*}
where the constants $w_0$ and $w_1$ were defined in \eqref{eq:w_0} and \eqref{eq:w_1}, respectively.

From Lemma \ref{lem:eigvals} and \eqref{eq:14} we have
\begin{align*}
\lamin_{\lfloor \sqrt{T_B} \rfloor + 1}) \geq \gamma - \ru + \frac{1}{d}\sum_{t=1}^{\lfloor \sqrt{T_B} \rfloor} \omega_t(\abam) \geq \frac{\gamma}{d}\left(u_1\log T_B - u_0\right) + \gamma - \ru.
\end{align*}
So
\begin{align*}
    \sum_{t=1}^{T_B} \omega_t(\abam) = & \sum_{t=1}^{\lfloor \sqrt{T_B} \rfloor}  \omega_t(\abam)  +  \sum_{t=\lfloor \sqrt{T_B} \rfloor + 1}^{T_B} \omega_t(\abam) \\
   \geq & \gamma(u_1 \log T_B - u_0) + \left(\frac{\gamma}{d}(u_1 \log T_B - u_0) + \gamma - \ru\right) \left(w_1 \log T_B - w_0\right)  \\
   = & \frac{\gamma}{d} u_1 w_1 (\log T_B)^2 + \left(\gamma u_1 - \frac{\gamma}{d}u_1w_0 - \frac{\gamma}{d} u_0 w_1 + \gamma w_1 - \ru w_1 \right)\log T_B \\ 
   &+ w_0\left(\frac{\gamma}{d} u_0 - \gamma + \ru\right) - \gamma u_0 \\
   = & d b_2(\log T_B)^2 + db_1 \log T_B + d(b_0 - \gamma + \ru),
\end{align*}
where the constants $b_0, b_1$ and $b_2$ were defined in \eqref{eq:b_0}, \eqref{eq:b_1}, and \eqref{eq:b_2}, respectively.

Then, applying Lemma \ref{lem:eigvals} again we get
\begin{align}\label{eq:lamin_lower_bound}
    \lamin_{T_B + 1}) \geq \gamma - \ru +\frac{1}{d} \sum_{t=1}^{T_B} \omega_t(\abam) \geq b_2 (\log  T_B)^2 + b_1 \log T_B + b_0.
\end{align}
Note that by Assumption \ref{as:alpha}, we have $u_1 > 0$ and $w_1 > 0$, so $b_2 > 0$.

From \eqref{eq:bound_log_det_frac} we have 
 \begin{align*}
     &{\beta}_{T_B + 1, \zeta_{T_B + 1}(\delta)}(\widehat{B}_t) = \left(\sqrt{2 \log (1 / \zeta_{T_B + 1}(\delta)) + \log \left(\frac{\det \bo{W}_{T_B + 1}}{\det \bo{W}_1}\right)} + \sqrt{\gamma} \widehat{B}_{T_B + 1}\right)^2 \leq \\
     & \leq 4 \log (1/\zeta_{T_B + 1}(\delta)) + 2 \log \left(\frac{\det \bo{W}_{T_B + 1}}{\det \bo{W}_1}\right) + 2\gamma \widehat{B}^2_{T_B + 1} \leq \\
     & \leq 4 \max\{\log(1/\delta), 2\log (T_B + 1)\} + 2d\log T_B + 2d \log \left(1 + \frac{\ru}{\gamma}\right) + 2\gamma B^2.
 \end{align*}
 Since $T_B \geq 4$ we have
$$
\log(T_B + 1) \leq \log\left(\frac{5}{4}T_B\right) = \log T_B + \log(5/4),
$$
so
\begin{align}\label{eq:20}
    {\beta}_{T_B + 1, \zeta_{T_B + 1}(\delta)}(\widehat{B}_t)
    \leq & (2d+8)\log T_B + 8\log (5/4) + 4\log(1/\delta) +2d\log\left(1+\frac{\ru}{\gamma}\right) + 2\gamma B^2 \nonumber \\
    = & (2d+8)\log T_B + h_0,
\end{align}
with $h_0$ defined in \eqref{eq:h_0}.
Note that $h_0  > 0$. Also, since $T_B \geq 4$ we have $\log T_B > 1$ so from \eqref{eq:lamin_lower_bound}, \eqref{eq:20} and the fact that $b_2 > 0$ we get
\begin{align*}
\frac{\lamin_{T_B + 1})}{{\beta}_{T_B + 1, \zeta_{T_B + 1}(\delta)}(\widehat{B}_t)} \geq & \frac{b_2(\log T_B)^2 + b_1 \log T_B + b_0}{(2d+8)\log T_B + h_0} \\
= & \frac{b_2}{2d+8+\frac{h_0}{\log T_B}} \log T_B + \frac{b_1}{(2d+8) + \frac{h_0}{\log T_B}} + \frac{b_0}{(2d +8)\log T_B + h_0} \\
\geq & \frac{b_2}{h_0 + 2d + 8} \log T_B - \frac{|b_1|}{2d +8} - \frac{|b_0|}{h_0 + 2d + 8}.
\end{align*}
 Note that by Assumption \ref{as:T_B} we have
 \begin{equation*}
     T_B \geq \exp\left[\frac{h_0 + 2d + 8}{b_2}\left(4g^{-2}B^{* -2} + \frac{|b_1|}{2d +8} + \frac{|b_0|}{h_0 + 2d + 8}\right)\right]
 \end{equation*}
 
 so
 \begin{equation*}
     \frac{\lamin_{T_B + 1})}{{\beta}_{T_B + 1, \zeta_{T_B + 1}(\delta)}(\widehat{B}_t)} \geq 4g^{-2}B^{* -2}
 \end{equation*}
 which is the required contradiction to \eqref{eq:12}. 
 So there exists $t \leq T_B + 1$ such that 
  \begin{equation*}
     \frac{\lamin_{t})}{{\beta}_{t, \zeta_{t}(\delta)}(\widehat{B}_t)} \geq 4g^{-2}B^{* -2}
 \end{equation*}
and thus, since $E_{\delta}$ holds, from \eqref{eq:11} for any $t \geq T_B + 1$ we have
\begin{equation*}
    B^* \leq \tilde{B}_t \leq (1+g)B^*. \qed
\end{equation*}

\subsection{Proof of Proposition \ref{prop:regret_2}}\label{proof_regret_2}
The exact assumptions made by Propositions \ref{prop:regret_2} are as follows. We assume that $T_B$ and $\alpha$ are sufficiently large so Assumptions \ref{as:B} - \ref{as:alpha} hold and $(T_B+1)^2 \geq 1/\delta$.
We can now proceed to the proof.

\proof 
Suppose that event $E_{\delta}$ holds.
\begin{align*}
    \E \left[ \sum_{t=1}^T \widehat{\Delta}_{t, \zeta_t(\delta)}(A_t^{\tn{BEIDS}})\right] = \E \left[ \sum_{t=1}^{T_B} \widehat{\Delta}_{t, \zeta_t(\delta)}(A_t^{\tn{BAM}})\right] + \E \left[ \sum_{t=T_B+1}^{T} \widehat{\Delta}_{t, \zeta_t(\delta)}(A_t^{\tn{EB-UCB}})\right].
\end{align*}

From \eqref{eq:bound_UCB_info} with probability $1$ we have
\begin{align}\label{eq:bound_EB_UCB_info_phase_2}
    \sum_{t=T_B+1}^{T} I_t^{\tn{EB-UCB}}(A_t) \leq O(d \log T).
\end{align}
Let $a_t^{\tn{EB-UCB}}$ be the realization of $A_t^{\tn{EB-UCB}}$. Since event $E_{\delta}$ holds and Assumptions \ref{as:B} - \ref{as:alpha} hold, from Proposition \ref{prop:high_prob_B} we have $B^* \leq \tilde{B}_t \leq (1+g)B^*$ for all $t \geq T_B + 1$. Also from the assumptions of this proposition, 
$2\log T \geq \log(1/\delta)$, so analogously as in \eqref{eq:Psi_bound_1} and \eqref{eq:bound_beta_T}, for any $t \in \{T_B+1, T_B+2, \ldots, T \}$ we have
\begin{align*}
    \frac{\widehat{\Delta}^2_{t,\zeta_t(\delta)}\left(a_t^{\tn{EB-UCB}}\right)}{I_t^{\tn{EB-UCB}}\left(a_t^{\tn{EB-UCB}}\right)} \leq & 16 {\beta}_{T, \zeta_T(\delta)}(\tilde{B}_t) \max \{U^2 \gamma^{-1}, \rho_{\max}^2\} \\
    \leq & 16 \max \{U^2 \gamma^{-1}, \rho_{\max}^2\} \\
    & \times \left[2\max \{ 2\log T, \log(1/\delta) \} +  2 d \log(T-1) +2 d \log\left(1 + \frac{\ru}{\gamma}\right) +  2\gamma \tilde{B}_t^2\right] \\
    \leq & 16 \max \{U^2 \gamma^{-1}, \rho_{\max}^2\} \\ 
    & \times \left[ (2 d + 4) \log T +2 d \log\left(1 + \frac{\ru}{\gamma}\right) +  2\gamma \tilde{B}_t^2\right] \\
   \leq & 16 \max \{U^2 \gamma^{-1}, \rho_{\max}^2\} \\ 
   & \times \left[ (2 d + 4) \log T +2 d \log\left(1 + \frac{\ru}{\gamma}\right) +  2\gamma \left((1+g)B^*\right)^2\right].
\end{align*}
Hence from Theorem \ref{thm:univ_IDS_bound}  and \eqref{eq:bound_EB_UCB_info_phase_2} we have
\begin{align*}
    \E \left[ \sum_{t=T_B+1}^{T} \widehat{\Delta}_{t, \zeta_t(\delta)}(A_t^{\tn{EB-UCB}})\right] \leq & O\left(\vphantom{\log\left(1 + \frac{\ru}{\gamma}\right)}{d \max \{U / \sqrt{\gamma}, \rho_{\max}\}} \sqrt{T} \log T \right. \\
    &\times \left . \sqrt{ \log\left(1 + \frac{\ru}{\gamma}\right) + \gamma \left((1+g)B^*\right)^2 }  \right) \\
    \leq & O\left(d U \rho_{\max}(1+g)B^* \sqrt{T} \log T\right),
\end{align*}
and thus from \eqref{eq:delta_hat_gap} we get that 
\begin{align*}
    \E \left[ \sum_{t=T_B+1}^{T} {\Delta}(A_t^{\tn{EB-UCB}})\right]
    \leq O\left(d U \rho_{\max}(1+g)B^* \sqrt{T} \log T\right).
\end{align*}
and similarly with probability $1$ we have
\begin{align*}
\sum_{t=T_B+1}^{T} {\Delta}(A_t^{\tn{EB-UCB}}) \leq O\left(d U \rho_{\max}(1+g)B^* \sqrt{T} \log T\right).
\end{align*}
Thus, since $T_B$ is fixed with respect to $T$ with probability at least $\Prob(E_{\delta}) \geq 1 -\delta$ we have
\begin{equation*}
    \mathcal{R}_T \leq O\left(d U \rho_{\max}(1+g)B^* \sqrt{T} \log T\right)
\end{equation*}
and 
\begin{equation*}
    \mathcal{PR}_T \leq O\left(d U \rho_{\max}(1+g)B^* \sqrt{T} \log T\right). \qed
\end{equation*}

\section{Additional simulation studies}

\begin{figure}[htbp]
    \centering
    \begin{subfigure}[b]{0.41\textwidth}
        \includegraphics[scale=0.29]{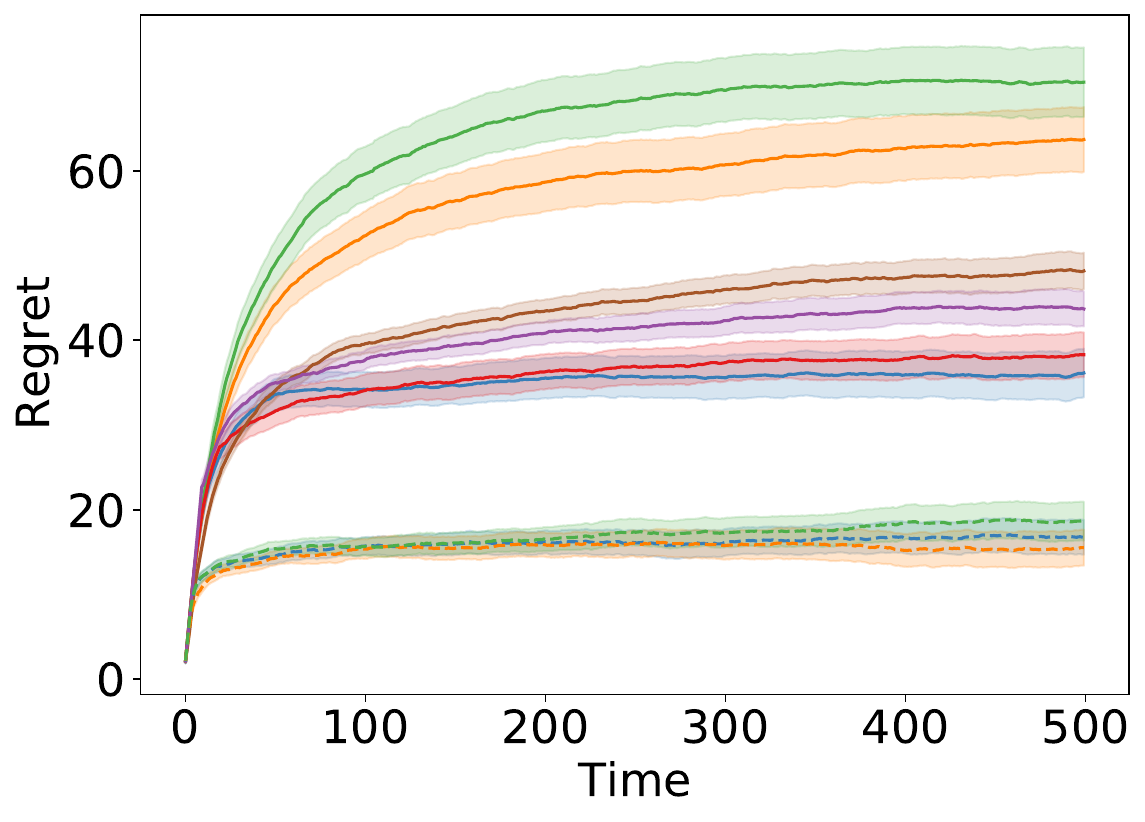}
        \caption{$10$ arms with randomized variances}
        \label{fig:10_arms_random}
    \end{subfigure}
    \hfill
    \begin{subfigure}[b]{0.58\textwidth}
        \includegraphics[scale=0.29]{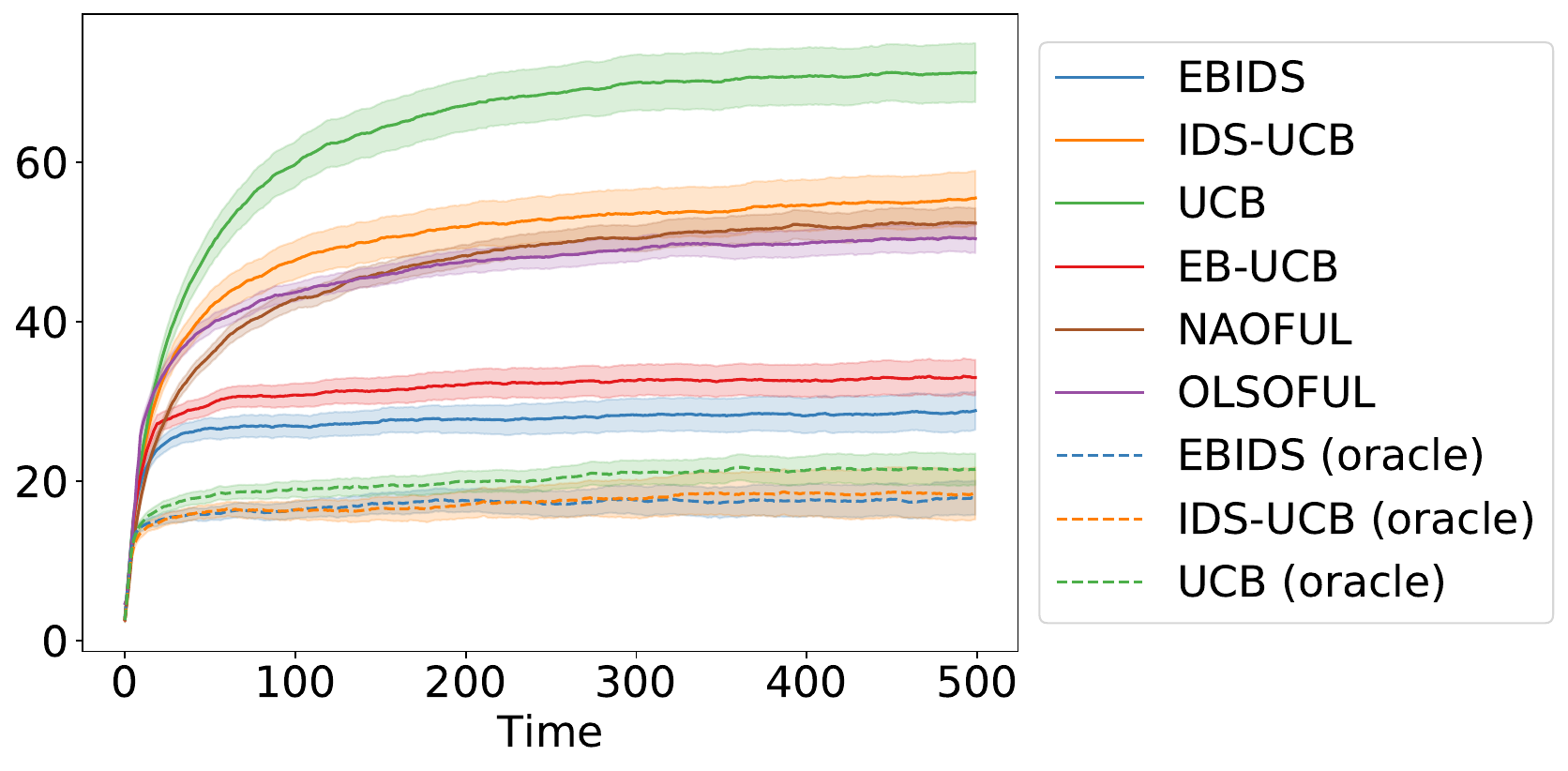}
        \caption{$20$ arms with randomized variances \phantom{aaaaaaaaaaaaaaaa}}
        \label{fig:20_arms_random}
    \end{subfigure}
    \begin{subfigure}[b]{0.40\textwidth}
        \includegraphics[scale=0.29]{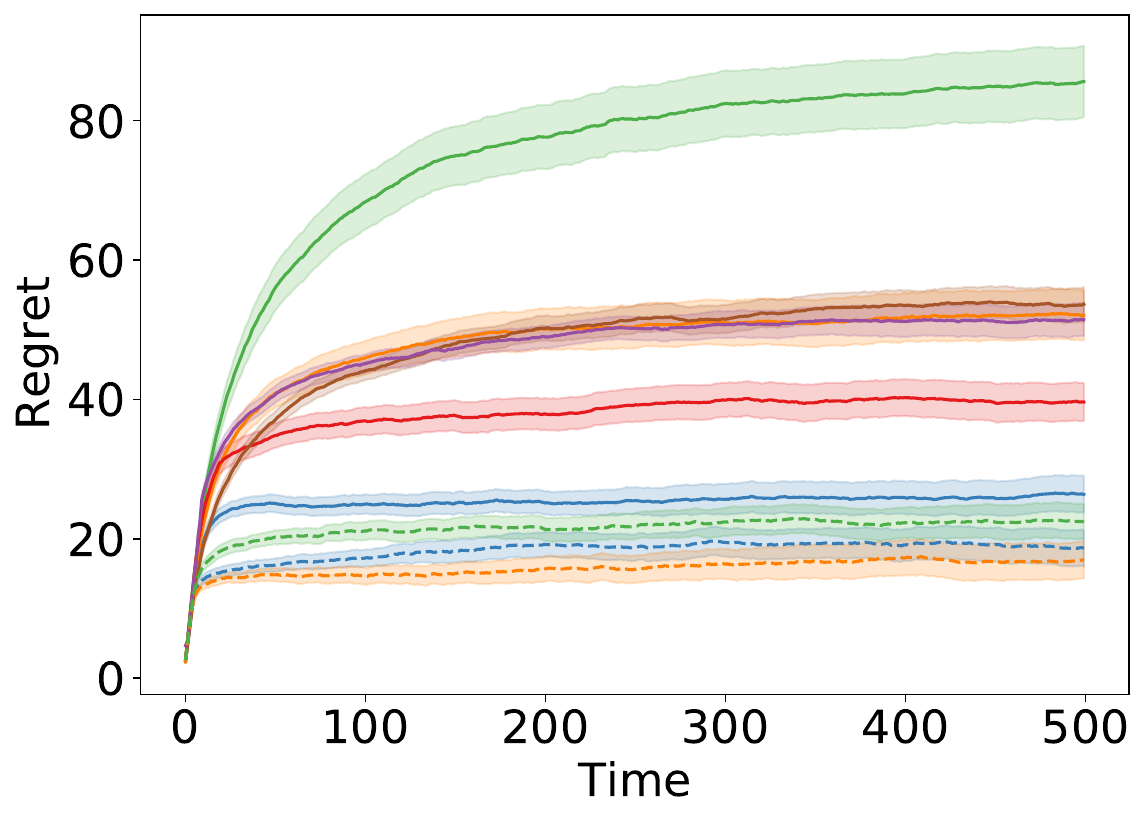}
        \caption{$20$ arms with fixed variances}
        \label{fig:20_arms_fixed}
    \end{subfigure}
    \hfill
    \begin{subfigure}[b]{0.59\textwidth}
        \includegraphics[scale=0.29]{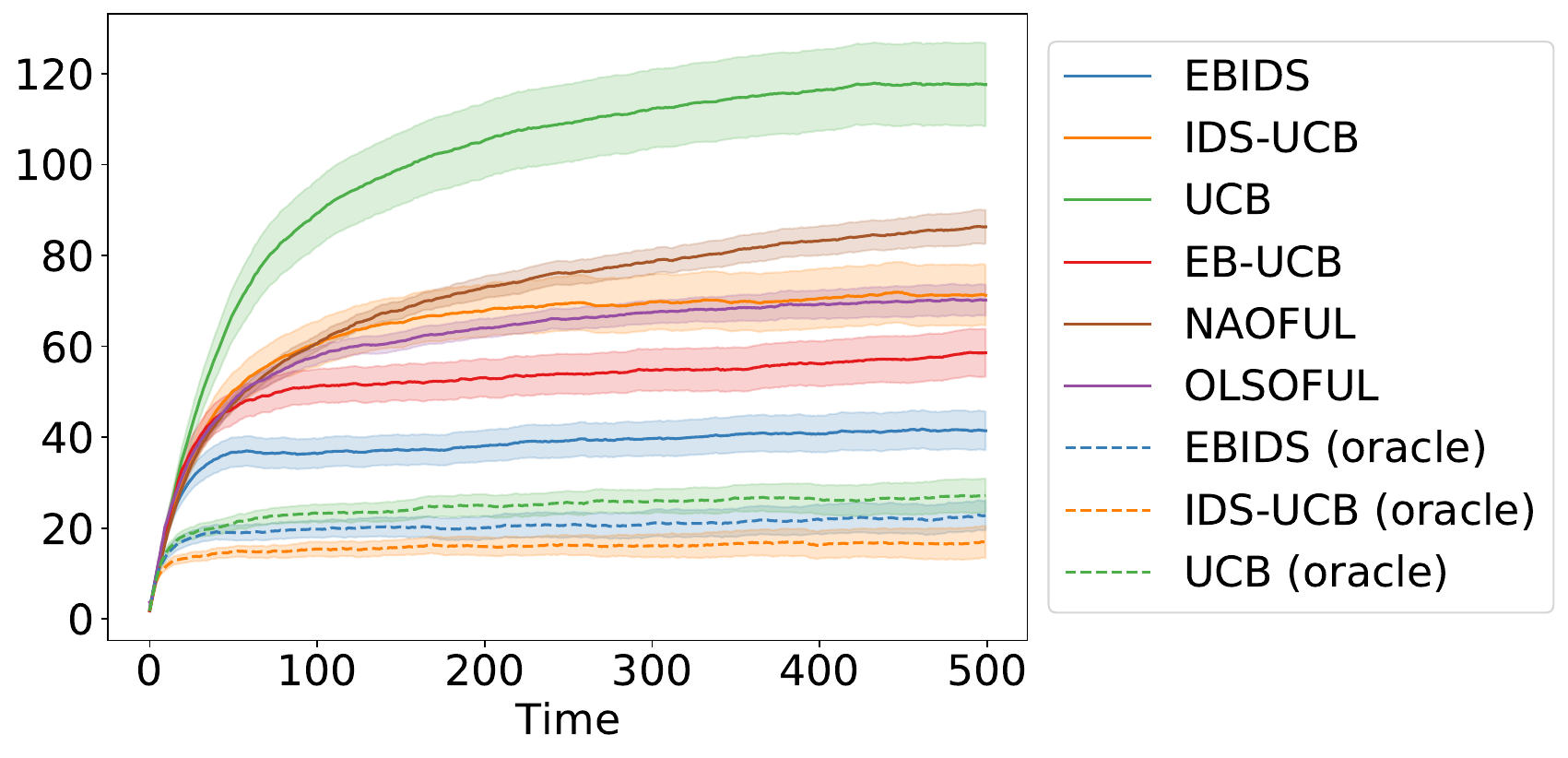}
        \caption{Continuous action space \phantom{aaaaaaaaaaaaaaaa}}
        \label{fig:splines}
    \end{subfigure}
    \caption{Regret incurred by EBIDS, EB-UCB, NAOFUL, OLSOFUL, IDS-UCB and UCB using conservative $B=100$ for simulation settings (a)-(d) outlined above. We include the oracle versions of EBIDS, IDS-UCB, and UCB using $B=B^*$ for reference. The solid and dashes lines represent the regret averaged over $200$ repeated experiments, while the shaded bounds are $95\%$ pointwise confidence bands. }
    \label{fig:additional_simulations}
\end{figure}

In this section we provide the results of additional simulation studies we ran. Similarly as above, we assume that the random noise terms $\eta_t$ are drawn from mean-zero normal distributions and $\bo{\theta}^* = [-5, 1, 1, 1.5, 2]^{\top}$ is the true parameter vector. We use the conservative $B=100$ upper bound for $\| \bo{\theta}^*\|_2$. We consider the following scenarios:
\begin{enumerate}
    \item[] (a) Ten arms where for each experiment the arm features are drawn independently from Unif$[-1/\sqrt{5}, 1/\sqrt{5}]$ and the standard deviations for these arms are drawn independently from Unif$[0.1, 1]$.
    \item[] (b) The same setup as (a), but with twenty arms instead of ten.
    \item[] (c) Twenty arms where for each experiment the arm features are drawn independently from Unif$[-1/\sqrt{5}, 1/\sqrt{5}]$, with ten arms yielding rewards with standard deviation $0.2$ and the other ten yielding rewards with standard deviation $1$.
    \item[] (d) A continuum of actions $\A = [0,1]$ where for each experiment the $k$-th coordinate $[\bo{\phi}(\cdot)]_k$ is drawn independently from the space of cubic B-splines with ten equally spaced knots, for any $k \in \{ 1, \ldots , 5\}$. The standard deviation of each arm $a \in \A$ is given by $\exp(0.5 - 3a)$. To implement this simulation we use a discretization of $\A = [0,1]$ into $1000$ equally spaced points.
\end{enumerate}
We present the results for settings (a) - (d) in Figure \ref{fig:additional_simulations} with regret averaged over $200$ repeated experiments of $T=500$ steps, along with $95\%$ pointwise confidence bounds. As we can see, across all the considered cases, EBIDS remains the best-performing algorithm among methods that do not have access to the true value of $\|\bo{\theta}^*\|_2$.
In setting (a), where the standard deviations for the ten arms are drawn from Unif$[0.1, 1]$ across experiments, EB-UCB becomes competitive with EBIDS. In general, the optimistic algorithms (UCB, EB-UCB, NAOFUL, OLSOFUL) perform comparatively better in this setting than they do in simulations where arm variances are fixed across the experiments presented in Figures \ref{fig:algos_comparison} and \ref{fig:20_arms_fixed}.
This is likely because, on average, the experiments in setting (a) involve fewer arms with very low variances, which reduces the advantage of IDS algorithms stemming from utilizing those highly informative arms. For the same reason, settings with larger numbers of arms tend to favor the IDS algorithms, as they provide more low-variance arms to exploit for information gain. This is evident from the improved performance of both EBIDS and IDS-UCB relative to the other algorithms in settings with twenty arms and the continuous action space, with EBIDS performance approaching even that of the oracle version of UCB in the former case.

\end{document}